\documentclass{article}

\usepackage[final]{neurips_2022}

\usepackage[utf8]{inputenc} 
\usepackage[T1]{fontenc}    
\usepackage{hyperref}       
\usepackage{url}            
\usepackage{booktabs}       
\usepackage{amsfonts}       
\usepackage{nicefrac}       
\usepackage{microtype}      
\usepackage{xcolor}         
\usepackage{graphicx}
\usepackage{wrapfig}
\usepackage{mathtools}
\usepackage{floatflt}
\usepackage[ruled,vlined,linesnumbered]{algorithm2e}
\title{A composable machine-learning approach for steady-state simulations on high-resolution grids}

\author{%
  Rishikesh Ranade
    \\
  Office of CTO\\
  Ansys Inc.\\
  Canonsburg, PA 15317 \\
  \texttt{rishikesh.ranade@ansys.com} \\
  \And
  Chris Hill \\
  Fluids Business Unit \\
  Ansys Inc. \\
  Lebanon, NH, 03766 \\
  \texttt{chris.hill@ansys.com} \\
  \AND
  Lalit Ghule \\
  Office of CTO \\
  Ansys Inc. \\
  Canonsburg, PA, 15317 \\
  \texttt{lalit.ghule@ansys.com} \\
  \And
  Jay Pathak \\
  Office of CTO \\
  Ansys Inc. \\
  San Jose, CA, 95134 \\
  \texttt{jay.pathak@ansys.com} \\
}

\begin{document}

\maketitle

\begin{abstract}
   In this paper we show that our Machine Learning (ML) approach, CoMLSim (\textbf{Co}mposable \textbf{M}achine \textbf{L}earning \textbf{Sim}ulator), can  simulate PDEs on highly-resolved grids with higher accuracy and generalization to out-of-distribution source terms and geometries than traditional ML baselines. Our unique approach combines key principles of traditional PDE solvers with local-learning and low-dimensional manifold techniques to iteratively simulate PDEs on large computational domains. The proposed approach is validated on more than $5$ steady-state PDEs across different PDE conditions on highly-resolved grids and comparisons are made with the commercial solver, Ansys Fluent as well as $4$ other state-of-the-art ML methods. The numerical experiments show that our approach outperforms ML baselines in terms of 1) accuracy across quantitative metrics and 2) generalization to out-of-distribution conditions as well as domain sizes. Additionally, we provide results for a large number of ablations experiments conducted to highlight components of our approach that strongly influence the results. We conclude that our local-learning and iterative-inferencing approach reduces the challenge of generalization that most ML models face.
\end{abstract}

\section{Introduction}

Engineering simulations utilize solutions of partial differential equations (PDEs) to model various complex physical processes like weather prediction, combustion in a car engine, thermal cooling on electronic chips etc., where the complexity in physics is driven by several factors such as Reynolds number, heat source, geometry, etc. Traditional PDE solvers use discretization techniques to approximate PDEs on discrete computational domains and combine them with linear and non-linear equation solvers to compute PDE solutions. The PDE solutions are dependent on several conditions imposed on the PDE, such as geometry and boundary conditions of the computational domain as well as source terms such as heat generation, buoyancy etc. Moreover, many applications in engineering require highly-resolved computational domains to accurately approximate the PDEs solutions as well as to capture the intricacies of conditions such as geometry and source terms imposed on the PDEs. As a result, traditional PDE solvers, although accurate and generalizable across various PDE conditions, can be computationally slow, especially in the presence of complicated physics and large computational domains. In our work, we specifically aim to increase the speed of engineering simulations using ML techniques on complex PDE conditions used for enterprise simulations.

\begin{figure}[t!]
\centering
\includegraphics[width=0.55\textwidth]{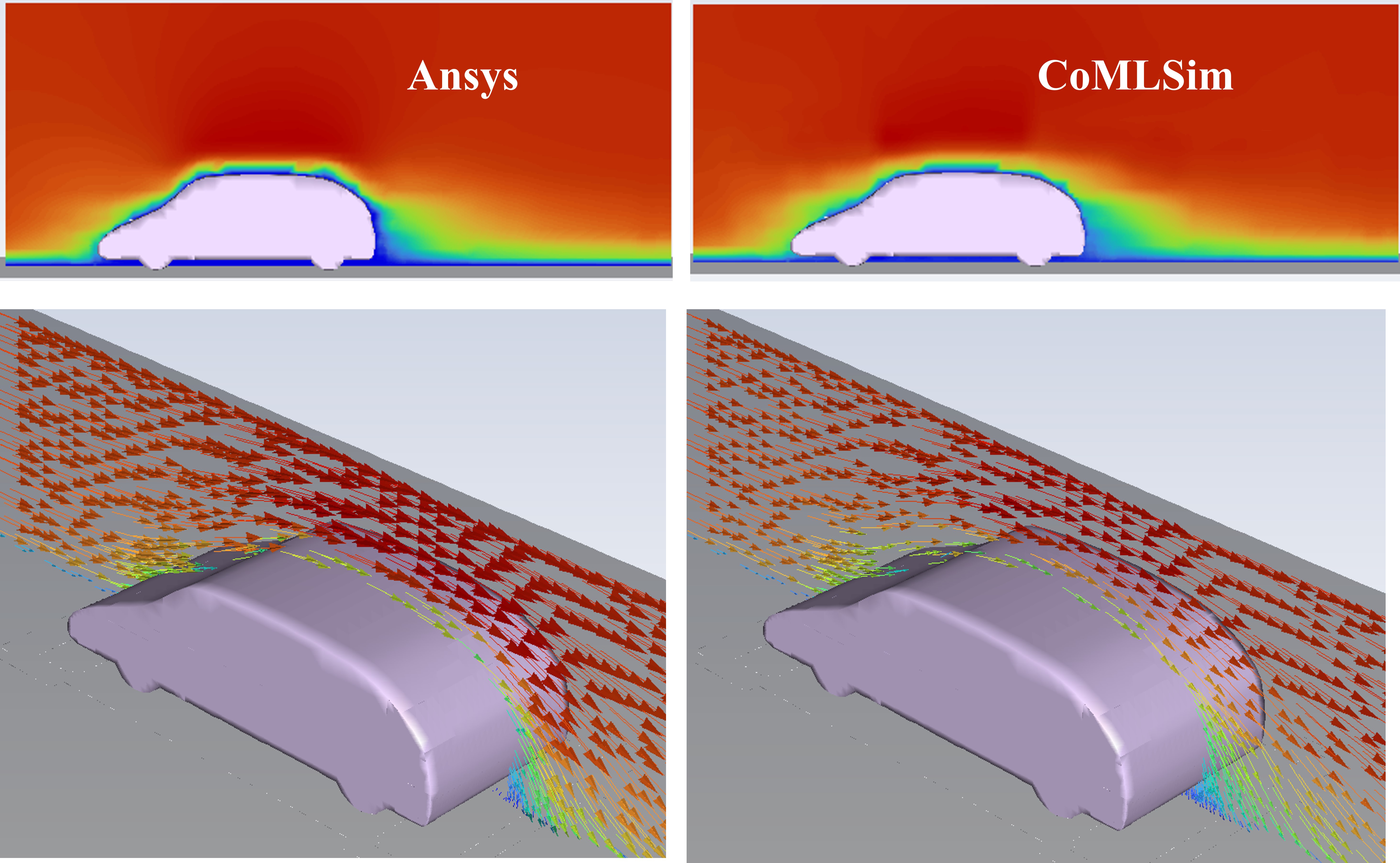}
\caption{CoMLSim performance vs Ansys Fluent for a high Reynolds number, steady-state turbulent flow over a car (never seen during training) at 90 mph. The vectors and contours predictions match reasonably with MAE of 0.0145. High flow gradients that affect the wake pattern are resolved accurately and comparable to Fluent.}
\label{fig:0}
\end{figure}

The idea of using Machine Learning (ML) with PDEs is not unique and has been explored for several decades \cite{crutchfield1987equations, kevrekidis2003equation}. ML approaches are computationally fast, but they fall short in terms of accuracy, generalization to a wide range and out-of-distribution PDE conditions and in their ability to scale to highly-resolved computational grids, when compared to traditional PDE solvers. A few shortcomings of the current methods are outlined below and subsequently are motivations for this work.

\begin{figure}[b]
\centering
\includegraphics[width=0.9\textwidth]{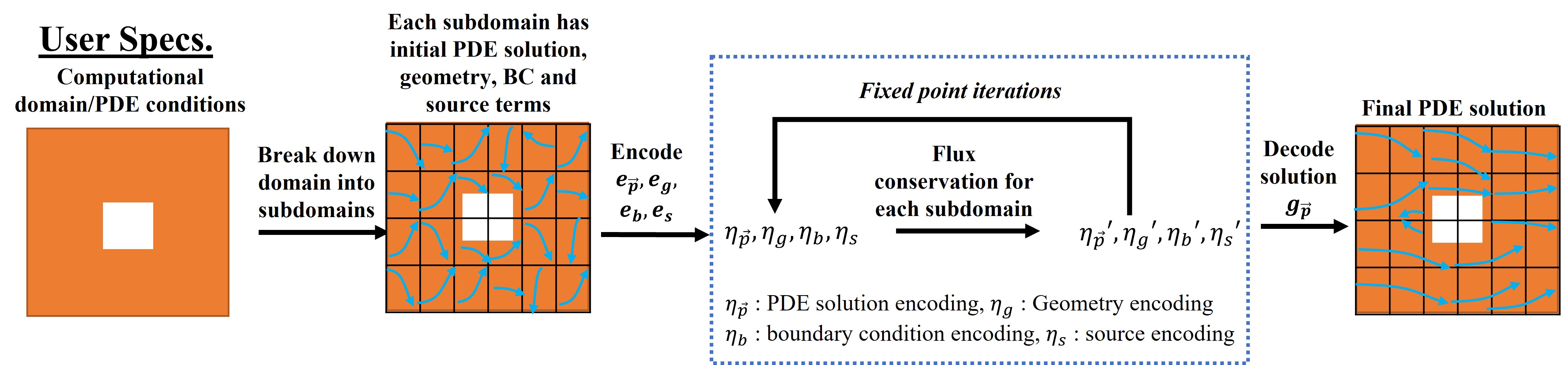}
\caption{CoMLSim solution algorithm}
\label{fig:1}
\end{figure}

\begin{enumerate}
    \item ML approaches use static inferencing to predict PDE solutions as a function of the PDE conditions. In many cases, solutions and conditions have high-dimensional and sparse representations, which are challenging to generalize with such black-boxed static inferencing.
    \item Most engineering applications require highly-resolved computational grids to capture detailed solution features, such as hot spots on an electronic chip surface. This adversely impacts training in terms of GPU memory, computational time and data requirements. 
    \item Most ML approaches fail to use powerful information from traditional PDE solvers related to solver methods, numerical discretization etc.
\end{enumerate}
In this work, we introduce a novel ML approach, \textbf{Co}mposable \textbf{M}achine \textbf{L}earning \textbf{Sim}ulator (CoMLSim) to work with high-resolution grids without compromising accuracy. To achieve that, our method decomposes highly-resolved computational grids into smaller local subdomains \color{black} with Cartesian grids\color{black}. PDE solutions and conditions on each subdomain are represented by lower-dimensional latent vectors determined using autoencoders. The latent vectors of solutions corresponding to conditions are predicted in each subdomain and stitched together to maintain local consistency, analogous to flux conservation in traditional solvers. In this paper, we describe our method (shown in fig. \ref{fig:1}) for building CoMLSim for different PDEs, showcase different experiments to compare accuracy and generalizability with different ML methods and provide an extensive ablation study to understand the impact of CoMLSim's different components on the results.  

\textbf{Significant contributions of this work:} 
\begin{enumerate}
    \item The CoMLSim approach combines traditional PDE solver strategies such as domain discretization, flux conservation, solution methodologies etc. with ML techniques to accurately model numerical simulations on high-resolution grids.
    \item Our approach operates on local subdomains and solves PDEs in a low-dimensional space. This enables generalizing to out-of-distribution PDE conditions and scaling to bigger domains with large mesh sizes. 
    \item The iterative inferencing algorithm is self-supervised and allows for coupling with traditional PDE solvers.
\end{enumerate}
\textbf{Related Works:}
The use of ML for solving PDEs has gained tremendous traction in the past few years. Much of the research has focused on improving neural network architectures and optimization techniques to enable generalizable and accurate learning of PDE solutions. More recently, there has been a lot of focus on learning PDE operators with neural networks (NNs) \cite{bhattacharya2020model, anandkumar2020neural, li2020multipole, patel2021physics, lu2021learning, li2020fourier, li2021physics}. The neural operators are trained on high-fidelity solutions generated by traditional PDE solvers on a computational grid of specific resolution and do not require any knowledge of the PDE. However, accuracy and out-of-distribution generalizability deteriorates as the resolution of computational grid increases (for example, $2048^2$ in 2-D or $256^3$ in 3-D). Furthermore, these methods are limited by an upper cap on GPU memory. 

A different research direction focuses on training neural networks with physics constrained optimization \cite{raissi2019physics, raissi2018hidden}. These method use Automatic Differentiation (AD) \citep{baydin2018automatic} to compute PDE derivatives. The physics-based approaches have been extended to solve complicated PDEs representing complex physics \citep{jin2021nsfnets, mao2020physics, rao2020physics, wu2018physics, qian2020lift, dwivedi2021distributed, haghighat2021physics, haghighat2021sciann, nabian2021efficient, kharazmi2021hp, cai2021flow, cai2021physics, bode2021using, taghizadeh2021explicit, lu2021deepxde, shukla2021parallel, hennigh2020nvidia, li2021kohn}. More recently, alternate approaches that use discretization techniques using higher order derivatives and specialize numerical schemes to compute derivatives have shown to provide better regularization for faster convergence \citep{ranade2021discretizationnet, gao2021phygeonet, wandel2020learning, he2020unsupervised}. However, the use of optimization techniques to solve PDEs, although accurate, has proved to be extremely slow as compared to traditional solvers and hence, non-scalable to high resolution meshes.

Domain decomposition techniques have been successfully used in the context of ML applied to PDEs \cite{heinlein2021combining}. These techniques enable local learning from smaller restricted domains and have proved to accelerate learning of neural networks. In \citep{heinlein2019machine, heinlein2021combining}, the authors proposed a hybrid method combining ML with domain decomposition to reduce the computational cost of finite element solvers. Similarly, domain decomposition has been extensively used in the context of physics informed neural networks to reduce training costs by enabling distributed training on multiple GPUs \citep{li2019d3m, jagtap2020conservative}. \cite{lu2021one} and \cite{wang2021train} learn on localized domains but infer on larger computational domains using a stitching process. \cite{bar2019learning} and \cite{kochkov2021machine} learn coefficients of numerical discretization schemes from high fidelity data, which is sub-sampled on coarse grids. \cite{beatson2020learning} learn surrogate models for smaller components to allow for cheaper simulations. \cite{greenfeld2019learning} learn local prolongation operators from discretization matrices, to improve the rate of convergence of the multigrid linear solvers. \color{black}Alternatively, other methods such as \cite{pfaff2020learning, xu2021conditionally, harsch2021direct} make us of graph-based models to improve generalization by learning on local elements. In this work, we use a Cartesian grid to discretize each subdomain. However, GNNs \cite{alet2019graph, iakovlev2020learning, liu2021multi, li2022graph, chen2021graph} as well as FEM use other discretizations such as triangle or polygonal meshes. In future, these ideas can be used to handle unstructured meshes in subdomains.\color{black} Additionally, other ML methods improve the learning of ML models by compressing PDE solutions on to lower-dimensional manifolds. This has shown to improve accuracy and generalization capability of neural networks \citep{wiewel2020latent, maulik2020reduced, kim2019deep, murata2020nonlinear, fukami2020convolutional, ranade2021latent}. \color{black}Although, here we propose an iterative inferencing approach in the context of solving PDEs, it is inspired from the implicit deep learning approach used in a variety of machine learning tasks \cite{bai2019deep, rubanova2021constraint, remelli2020meshsdf, du2022learning}\color{black}. In this work, domain decomposition is combined with latent space learning is employed to accurately represent solutions on local subdomains and allow scalability to bigger computational domains. 


\section{CoMLSim approach} \label{comlsim_approach}

\subsection{Similarities with traditional PDE solvers} \label{similarities}

Consider a set of coupled steady-state PDEs with $n$ solution variables ($n=2$ for example). The coupled PDEs are defined as follows:
\begin{equation}
        L_1(u, v) - F_1  = 0; L_2(u, v) - F_2 = 0
    \label{eq1}
\end{equation}
$u(x,y,z)$ and $v(x,y,z)$ are defined on a computational domain $\Omega$ with boundary conditions specified on the boundary of the computational domain, $\Omega_b$. Here, $L_1,$ $L_2$ denote PDE operators, $F_1,$ $F_2$ represent PDE source terms \color{black}and $u = u_b, v = v_b$ for $\Omega=\Omega_b$\color{black}. The PDE operators can vary for different PDEs. For example, in a non-linear PDE such as the unsteady, incompressible Navier-Stokes equation the operator, $L = \vec{a}.\vec{\nabla} - \vec{\nabla}.\vec{\nabla}$ Traditional \color{black}FVM/FDM based \color{black} PDE solvers solve PDEs in Eq. \ref{eq1} by computing solutions variables, $u, v$, and their linear/non-linear derivatives on a discretized computational domain. Iterative solution algorithms are used to conserve fluxes between neighboring computational elements and determine consistent PDE solutions over the entire domain at convergence. The CoMLSim approach is designed to perform similar operations but at the level of subdomains ($n^3$ elements) using ML techniques. Additional details are provided in supplementary materials sections A.4.
 
\subsection{Solution algorithm for steady-state PDEs} \label{solution_algorithm}
The solution algorithm of CoMLSim approach is shown in Fig. \ref{fig:1} and described in detail in Alg. \ref{alg:predict}. Similar to traditional solvers, the CoMLSim approach discretizes the computational domain into smaller subdomains contain $n^p$ computational elements, where $p$ refers to spatial dimensionality and $n$ is predetermined. Each subdomain has a constant physical size and represents both PDE solutions and conditions. For example, a subdomain cutting across the cylinder in Fig. \ref{fig:1} represents the geometry and boundary conditions of the part-cylinder as well as the corresponding solution. In a steady-state problem, the solutions on the domain are initialized either with uniform solutions or they are generated from coarse-grid PDE solvers. The initial solution $\vec{p}=(u(\vec{x}), v(\vec{x}))$ is shown with randomly oriented flow vectors in Fig. \ref{fig:1}. Pretrained encoders $e_{\vec{p}}, e_g, e_b, e_s$ are used to encode the initial solutions as well as user-specified PDE conditions into lower-dimensional latent vectors, $\eta_{\vec{p}}$ corresponding to PDE solutions and $\eta_g, \eta_b, \eta_s$ corresponding to geometry, boundary conditions and source terms, respectively. Solution and condition latent vectors on groups of neighboring subdomains are concatenated together evaluated using a pre-trained flux conservation autoencoder ($\Theta$) to get a new set of solution latent vectors ($\eta'_{\vec{p}}$) on each subdomain, which are more locally consistent than the original vectors. The new solution latent vectors are iteratively passed through $\Theta$ to improve their local consistency and the iteration is stopped when the $L_2$ norm of change in solution latent vectors meets a specified tolerance, otherwise the iteration continues with the updated latent vectors. The latent vectors of the PDE conditions are not updated and help in steering the solution latent vectors to an equilibrium state that is decoded to PDE solutions using pretrained decoders $(g_{\vec{p}})$ on the computational domain. The converged solution in Fig. \ref{fig:1} is represented with flow vectors that are locally consistent with neighboring subdomains.The iterative procedure used in the CoMLSim approach can be implemented using several linear or non-linear equation solvers, such as Fixed point iterations \citep{bai2019deep}, Gauss Seidel, Newton's method etc., that are used in commercial PDE solvers. Out of those we have explored Point Jacobi and Gauss Seidel, which are described below. The use of autoencoders for compressed solution and condition representation in tandem with iterative inferencing solution algorithm is inspired from \cite{ranade2021latent,ranade2021discretizationnet} but the main difference in this work is that the solution procedure is carried out on local subdomains as opposed to entire computational domains to align with principles used in traditional PDE solvers. Supplementary material section G explains our algorithm in more details using an example.

\begin{algorithm}[H]
\caption{Solution methodology of CoMLSim approach}
\label{alg:predict}
Domain Decomposition: Computational domain $\Omega$ $\rightarrow$ Subdomains $\Omega_c$ \\

Initialize solution on all $\Omega_c$: ${\vec{p}}(\vec{x}) = 0.0$ for all $\vec{x} \in \Omega_c$

Encode solutions on $\Omega_c$: $\eta_{{\vec{p}}} = e_{u}({{\vec{p}}(\Omega_c)})$  

Encode conditions on $\Omega_c$: 

$\eta_g = e_g(g(\Omega_c))$, $\eta_b = e_b(b(\Omega_c))$, $\eta_s = e_s(s(\Omega_c))$

Set convergence tolerance: $\epsilon_t = 1e^{-8}$ \\
\While{$\epsilon$ $>$ $\epsilon_t$}{
  \For{$\Omega_c \in \Omega$}{
    \text{Gather neighbors of}  $\Omega_c$: $\Omega_{nb} = [\Omega_c, \Omega_{left}, \Omega_{right}, ...]$ \\
    $\eta_{{\vec{p}}}' = \Theta(\eta^{nb}_{{\vec{p}}}, \eta^{nb}_b, \eta^{nb}_g, \eta^{nb}_s)$
}
Compute $L_2$ norm: $\epsilon = ||\eta_{{\vec{p}}}-\eta_{{\vec{p}}}'||^2_2$\\
Update: $\eta_{{\vec{p}}} \leftarrow \eta_{{\vec{p}}}'$ \text{for all} $\Omega_c \in \Omega$
}
Decode PDE solution on all $\Omega_c$: ${\vec{p}} = g_{\vec{p}}(\eta_{{\vec{p}}\Omega_c)})$
\end{algorithm}

\textbf{Why flux conservation networks work?} The flux conservation network ($\Theta$) is simply an autoencoder which takes encoding of solutions and conditions on a group of neighboring subdomains as both inputs and outputs. Similar to traditional solvers that use approximations to represent relationships between neighboring solutions on elements, the flux conservation autoencoder does the same for encoded subdomain solutions and corresponding conditions. The training of this network is carried out on converged, locally consistent solutions generated for a specific PDE for arbitrary conditions. Since this network has only learnt locally consistent solutions, if one were to initialize a group of neighboring subdomains with random noise and iteratively pass it through this network the output of such a procedure would be an equilibrium solution corresponding to some locally consistent PDE solution. However, the correct PDE solution it converges to depends on the fixed condition specified in this procedure. Additional results and details are provided in the supplementary materials sections A.1, A.4 and C.1.5.

\textbf{Stability of flux conservation autoencoder:}
The iterative inferencing approach proposed in this work is similar to traditional approaches used in solving system of linear equations. Our algorithm uses fixed point iterations to solve Eq. \ref{eq2}
\begin{equation}
        \eta_{\vec{p}}' = \Theta(\eta_{\vec{p}}^{nb}, \eta_b^{nb}, \eta_g^{nb}, \eta_s^{nb})
    \label{eq2}
\end{equation}
where, $\Theta$ corresponds to the weights of flux conservation network, $\eta_{()}$ refers to the latent vectors and $nb$ refers to neighboring subdomains defined below in Eq. \ref{eq4}.
\begin{equation}
        \eta_{\vec{p}}^{nb} = \left[\eta_{\vec{p}}^s, \eta_{\vec{p}}^{left}, \eta_{\vec{p}}^{right}, \eta_{\vec{p}}^{top}, \eta_{\vec{p}}^{bottom}\right]
    \label{eq4}
\end{equation}

Similar to linear systems, the stability of our iterative inferencing approach is governed by the condition number \citep{trefethen1997numerical} defined in Eq. \ref{eq3}
\begin{equation}
        ||\frac{\partial \Theta(\eta_{\vec{p}}, \eta_b, \eta_g, \eta_s)}{\partial \eta_{\vec{p}}}|| < 1
    \label{eq3}
\end{equation}

\textbf{Solution algorithms: Point Jacobi vs Gauss Seidel:} \label{pjvsgs}
Since the algorithm loops over all the subdomains in a specified order, while updating the solution encoding of a subdomain $c$, there are subdomains in the neighborhood that already have an updated solution encoding. The Gauss Seidel method uses these updated solution encodings ($\eta_{\vec{p}}'$) on neighboring subdomains to update the solution encoding on subdomain, $c$. On the other hand, Point Jacobi method does not make use of this and hence can be easily vectorized for significantly faster computation. The implementation has a remarkable similarity with how traditional solvers use these methods \cite{saad2003iterative} and leaves the door open to use other non-linear optimizers with physics-based constraints.

\subsection{Neural network components in CoMLSim}
Our algorithm employs two types of autoencoders, CNN autoencoders to establish a lower-dimensional representation of PDE solutions and conditions from local subdomains grids and FCNN Autoencoders for flux conservation, where the goal is to learn a reduced representation of solution and condition latent vectors in a local neighborhood. Let us consider an example of solving the Laplace equation, $\nabla^2 u = 0$ in 2D for arbitrarily shaped computational domains, $\Omega_g$. In this example, the CoMLSim algorithm will require $3$ autoencoders, described in Eq. \ref{eq5}, to learn lower-dimensional representations on local subdomains.
\begin{equation}\label{eq5}
\left. \begin{array}{ll}  
 \quad\quad\quad\quad\quad u' = S(u, \eta_u)\quad\quad \quad\quad\quad\quad\quad\quad\quad\quad\\[2pt]
 \quad\quad\quad\quad\quad g' = G(g, \eta_g)\quad\quad\quad\quad\\[2pt]
 \quad\quad\quad\quad\eta_{nb}'= \Theta(\eta_{nb}, \zeta)\quad\quad \quad\quad\quad\quad\quad\quad\quad\\[2pt]
 \end{array}\right\}
\end{equation}
where, $u, \eta_u, S$ refers to the solution, its latent vector and the weights of the autoencoder, respectively,  $g, \eta_g, G$ refers to a representation of geometry, such as Signed Distance Fields (SDF) \citep{maleki2021geometry}, its latent vector and the weights of the autoencoder, $\Theta$ refers to the weights of flux conservation autoencoder and $\eta_{nb}$ represents a set of concatenated latent vectors, $\eta_u$ and $\eta_g$ on a group of neighboring subdomains. All the autoencoders are trained with samples of PDE solutions generated for the same use case. In the case of coupled PDEs, a single autoencoder is trained for all solution variables. Additional details related to the autoencoder networks are provided in the supplementary materials section A.

\textbf{Why Autoencoders?:}
Solutions to classical PDEs such as the Laplace equation can be represented by homogeneous solutions as follows:
\begin{equation}
        \phi(x,y) = a_0 + a_1x + a_2y + a_3(x^2-y^2) + a_4(2xy) + ...
    \label{eq6}
\end{equation}
where, $\vec{A} = {a_0, a_1, a_2, ..., a_n}$ are constant coefficients that can be used to reconstruct the PDE solution on any local subdomain. $\vec{A}$ can be considered as a compressed encoding of the Laplace solutions. Since, it is not possible to explicitly derive such compressed encodings for other high dimensional and non-linear PDEs, the CoMLSim approach relies on autoencoders to compute them. It is known that non-linear autoencoders with good compression ratios can learn powerful non-linear generalizations \citep{goodfellow2016deep, rumelhart1985learning, bank2020autoencoders}. Autoencoders also have great denoising abilities, which improve robustness and stability, when used in iterative settings \citep{ranade2021latent}. In their paper, \citet{park2020stability} demonstrate that the latent manifold established by a trained autoencoder is stable for varying intensities of Gaussian noise. This quality of autoencoders is useful for the convergence of the iterative inferencing approach. 

\section{Experiments}\label{expt}
In this section as well as in the supplementary materials section C, we consider a number of use cases with varying degrees of difficulty resulting from the PDE formulation as well as source terms, geometry and boundary conditions. The PDEs have applications in fluid mechanics, structural mechanics and semiconductor simulations.

\textbf{Details of experiments:} Here we provide some details about the experiments. Additional details and experiments on Laplace and Darcy equation may be found in the supplementary materials section D.
\begin{enumerate}
    \item \textbf{2D Poisson equation:} The Poisson's equation, shown in Eq. \ref{eq8}, is very popular in engineering simulations, for example, chip temperature prediction, pressure equation in fluids etc.
    \begin{equation}
            \nabla^2 u = f
        \label{eq8}
    \end{equation}
    where, $u$ is the solution variable and $f$ is the source term. This PDE is solved on a $1024$x$1024$ grid to resolve the high-frequency features of the source term. The source, $f$, is sampled from a Gaussian mixture model shown in Eq. \ref{eq9}.
    \begin{equation}
        \sum_{j=0}^{1024}\sum_{j=0}^{1024} f_{i,j} = \sum_{j=0}^{1024}\sum_{j=0}^{1024}\sum_{k=0}^{30} A_k\exp\left( -\left(\frac{x-\mu_{x,k}}{\sigma_{x,k}}\right)^2 - \left(\frac{y-\mu_{y,k}}{\sigma_{y,k}}\right)^2 \right)
      \label{eq9}
    \end{equation}  
    where, $x, y$ correspond to the grid coordinates. $A_k$ randomly assumes either $0$ or $1$ to vary the number of active Gaussians in the model. $\mu_x, \mu_y$ and $\sigma_x, \sigma_y$ are the mean and standard deviations of Gaussians in $x$ and $y$ directions, respectively. The means and standard deviations vary randomly between $0$ to $1$ and $0.001$ to $0.01$, respectively. The smaller magnitude of standard deviation results in hot spots that require highly-resolved grids. In this case, $256$ training and $100$ testing solutions are generated using Ansys Fluent.
    \item \textbf{2D non-linear coupled Poisson equation:} The coupled non-linear Poisson's equation is shown below in Eq. \ref{eq10}. $u, v$ are the solution variables and $f$ is the source term, similar to the description in Eq. \ref{eq9}. This PDE has applications in reactive flow simulations.
    \begin{equation}
        \nabla^2 u = f - u^2;  \quad\quad  \nabla^2 v = \frac{1}{u^2+\epsilon} - v^2
        \label{eq10}
    \end{equation}
     The data generation is similar to Experiment $2$ with a slight difference that the source term is less stiff with a standard deviation that varies between $0.005$ to $0.05$.
    \item \textbf{3D Reynolds-Averaged Navier-Stokes external flow:} This use case consists of a 3-D channel flow with resolution $304$x$64$x$64$ at high Reynolds number over arbitrarily shaped objects. The corresponding PDEs are presented in supplementary materials section C.3 or may be referred from \citep{chorin1968numerical}. The characteristics of flow generated on the downstream has important applications in the design of automobiles and airplanes. In this case, we use $5$ primitive 3-D geometries namely, cylinder, cuboid, trapezoid, airfoil wing and wedge and their random combinations and rotational augmentations to create $150$ training geometries and $50$ testing geometries, which are solved with Ansys Fluent to generate the data. Out-of-distribution testing is carried out on simplistic automobile geometries as shown in Figure \ref{fig:0}.
    \item \textbf{3D chip cooling with Natural convection:} In this case, we extend the complexity of Experiments $1, 2$ and $3$ to a industrial 3-D case of chip cooling with natural convection. In this problem, the computational grid ($128^3$ resolution) consists of a chip that is subjected to the powermap specified by Eq. \ref{eq9}. The heating of the chip results in generation of velocity in the fluid and at steady-state, a balance is achieved. We use Ansys Fluent to generate $300$ training and $100$ testing solutions for velocity, pressure, temperature for arbitrary sources.
\end{enumerate}

\section{Results and Ablation studies}\label{res_ans}
In this section, we provide a variety of results for the experiments outlined in Section \ref{expt}. Additionally, we conduct thorough ablation studies to understand the different components of the CoMLSim approach in more detail. We use the 2-D Linear Poisson's equation experiment for these studies unless otherwise specified. Details regarding the CoMLSim set up for each experiment, baseline network architectures as well as additional results corresponding to contour and line plots and comparisons across other metrics are provided in supplementary materials section C.
\subsection{Comparison with Ansys Fluent and other ML baselines} \label{ml_baseline}
The CoMLSim is compared with other ML baselines namely, UNet \citep{ronneberger2015u}, FNO \citep{li2020fourier} and DeepONet \cite{lu2021learning} for the experiments outlined in Section \ref{expt}. We compare the mean absolute error with respect to Ansys Fluent and averaged over the unseen testing cases and all solution variables. In the case of 3-D chip cooling we report the $L_{\infty}$ norm of temperature as this metric is more suited to this industrial application. It may be observed from Table \ref{tab:baseline} that the CoMLSim approach performs better than other ML baselines. All the ML methods perform better for the non-linear Poisson's case as compared to the linear Poisson's because the source term is less stiff but the CoMLSim approach does a better job in modeling the non-linearity arising due solution coupling. \color{black}It must be noted that CoMLSim as well as the baselines are trained with reasonably training samples. The baseline performance can have different outcomes with increasing the size of the training data.\color{black}

\begin{table}[h]
\centering
\setlength{\belowcaptionskip}{-1pt}
\caption{Comparison with baselines}
\begin{tabular}{ |p{4cm}||p{1cm}||p{1.5cm}|p{1cm}|p{1cm}|p{1.5cm}|p{1cm}| }
\hline
Experiment&Metric& CoMLSim & UNet & FNO & DeepONet&FCNN\\
\hline
2-D Linear Poisson's & $L_1$   & 0.011    &0.132&   0.031 & 0.061&0.267\\
2-D non linear Poisson's & $L_1$ &   0.0053  & 0.0877   &0.0278 & 0.527&0.172\\
3-D NS external flow & $L_1$ &   0.012  & 0.0625   &0.038 & 0.81 & 0.125\\
3-D chip cooling & $L_\infty$ &   15.2  & 95.21   & 60.836 & 45.27 & 192.7\\
\hline
\end{tabular}
\label{tab:baseline}
\end{table}
\subsection{Assessment of generalizability}
In this section, we assess the generalizability of our method specifically with respect to out-of-distribution PDE conditions and scaling to higher resolution grids. The results are compared with Ansys Fluent as well as other ML baselines discussed in section \ref{ml_baseline}.

\begin{figure}[t!]
\centering
\includegraphics[width=1\linewidth]{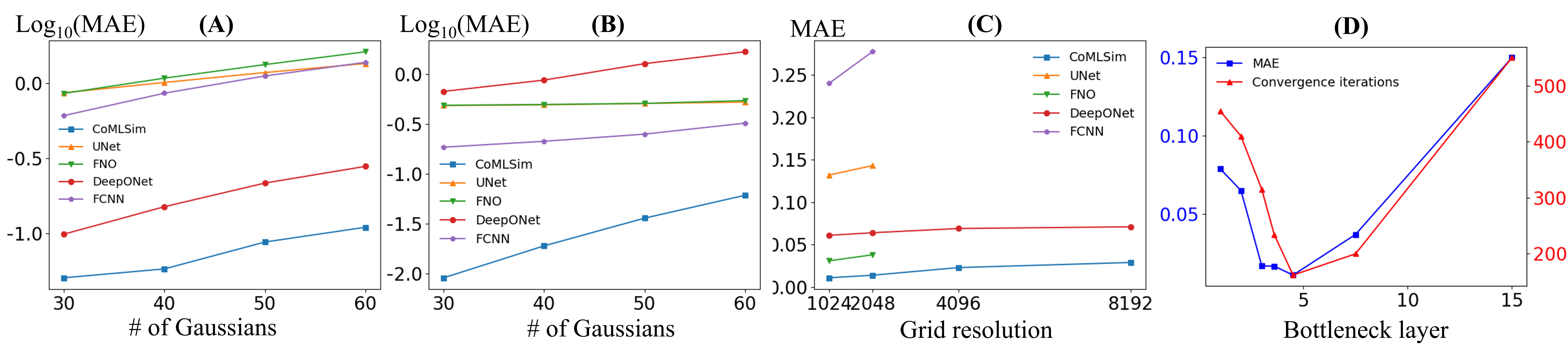}
\caption{(A): Error vs No. of Gaussians for linear Poisson's; (B): Error vs No. of Gaussians for non-linear Poisson's; (C): Error vs grid resolution for linear Poisson's; (D): MAE and convergence iterations vs flux conservation bottleneck size}
\label{fig:2}
\end{figure}

\textbf{Out-of-distribution source term:}
The source term for linear and non-linear Poisson equations are sampled from a Gaussian mixture model described in Eq. \ref{eq9}, where the number of Gaussian is randomly chosen between $0$ and $30$. In this experiment, we evaluate the generalizability of our approach for source terms with exactly $30$, $40$, $50$ and $60$ number of Gaussians, corresponding to out-of-distribution for the training data distributions. For each case, the results averaged over $10$ testing samples are compared with Ansys Fluent. 
\begin{wrapfigure}{r}{0.3\textwidth}
    \centering
    \vspace{-1em}
    \includegraphics[width=0.3\textwidth]{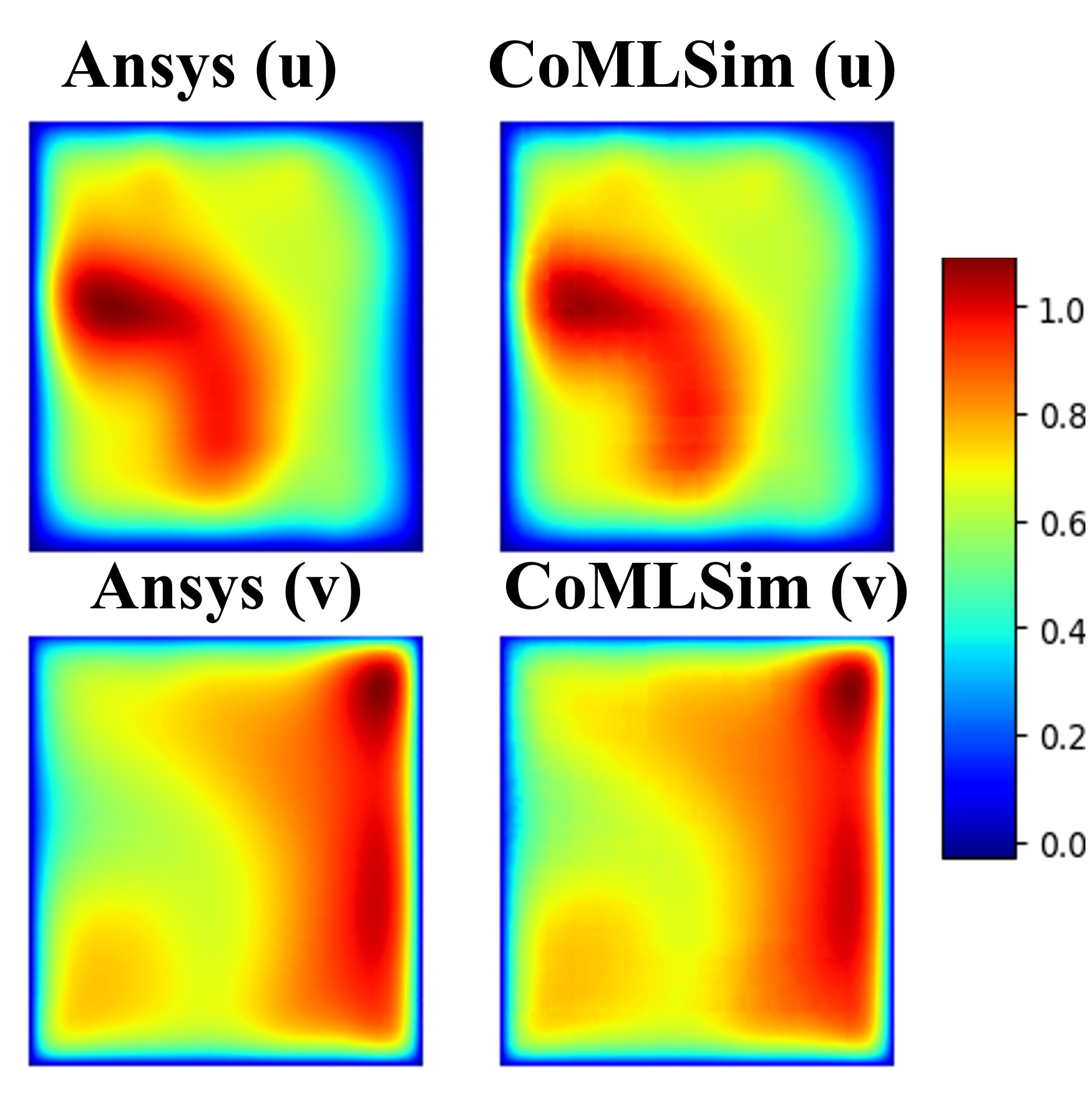}
    \vspace{-5ex}
    \caption{Example of out-of-distribution source (60 Gaussians) comparison for non-linear Poisson's}
    \label{fig:3}
\end{wrapfigure}
It may be observed from \ref{fig:2}A and B, that the accuracy of all ML approaches decreases as you move further away from the source term distribution. However, the CoMLSim approach performs significantly better than other ML baselines and the accuracy is reasonable even for the case with $60$ Gaussian mixture model, which is substantially different from the training distribution.



\textbf{Increasing mesh size by increasing domain size:}
Next, we evaluate the performance of CoMLSim for bigger domains with larger grid size and compare it with other ML baselines. In this experiment, we compute solutions for the $100$ source term conditions in the test set but on $4$ different domains with mesh sizes, $1024^2$, $2048^2$, $4096^2$, $8192^2$, respectively. 
The mean and standard deviation of Gaussians in the testing set are proportionally increased with the mesh size to ensure that the problem definition does not change. It is important to note that none of the networks are retrained for larger sized grids. It maybe observed from table \ref{fig:2} that the CoMLSim continues to scale to larger mesh sizes with similar accuracy as the original mesh size. The slight drop in accuracy due to increase in mesh size is attributed to accumulation of error caused by the flux conservation networks resulting from the increase in number of subdomains. 
Other ML baselines such as UNet and FNO scale up to a grid size of $2048^2$ but cannot evaluate beyond that due to GPU memory constraints. DeepONet has a point-wise inference and can also scale to larger-sized grids but its accuracy is lower than CoMLSim.

\textbf{Out-of-distribution geometries:} Next, we evaluate the performance of CoMLSim on $2$ unseen car models presented at different 3-D angles of rotation to the external flow. It should be noted that these geometries are more complicated than the primitive objects considered for training our approach \ref{expt}. The mean absolute errors with respect to Ansys Fluent across all solution variables are 0.0145, 0.06924, 0.04175, 0.91 and  0.13725 for CoMLSim, UNet, FNO, DeepONet and FCNN, respectively. Additional results and analysis are provided in supplementary materials C.3.3 and C.3.4. 



\subsection{Analysis of Subdomain size}\label{sub_size}
In this experiment, we analyze the effect of subdomain resolution on accuracy and computational cost of the CoMLSim approach. We train $3$ instances at resolutions of $32^2$, $64^2$ and $128^2$. The compression ratio in autoencoders is kept the same for the different subdomain resolutions. The mean absolute errors on testing set are $0.015$, $0.011$ and $0.029$, respectively. Additionally, the iterations required to convergence are $420$, $162$ and $89$, respectively. The convergence history is shown in Fig. \ref{fig:10}A.
The $128^2$ subdomain resolution has a lower accuracy because it is challenging to train accurate autoencoders as bigger subdomains capture a large amount of information.  On the other hand, the computational cost is the lowest for the $128^2$ resolution because the number of subdomains in the entire computational domain are significantly less and hence, the solution algorithm converges faster. 
\subsection{Flux conservation bottleneck layer size}
The flux conservation autoencoder as it is the primary workhorse of the CoMLSim solution algorithm. As a result, we evaluate the effect of the bottleneck layer size on the accuracy and computational speed. Each instance of CoMLSim with different bottleneck size is tested on $100$ unseen test cases. We verify that all models satisfy the stability criterion specified in Eq. \ref{eq4}. 
It may be observed from Figure \ref{fig:2}D that as the compression ratio of the flux conservation autoencoder decreases, it begins to overfit and the testing error as well as the number of convergence iterations and computational time significantly increase. On the other hand, if the compression ratio is too large the testing error increases because the autoencoders underfit. In alignment with the collective intuition about autoencoders, there exists an optimum bottleneck size compression ratio where the best testing error is obtained for small computational times.
\subsection{Robustness and stability} \label{robustness}
A long standing challenge in the field of numerical simulation is to guarantee the stability and convergence of non-linear PDE solvers. However, we believe that the denoising capability of autoencoders \citep{vincent2010stacked, goodfellow2016deep, du2016stacked, bengio2013generalized, ranzato2007efficient} used in our iterative solution algorithm presents a unique benefit, irrespective of the choice of initial conditions. Here, we empirically demonstrate the robustness and stability of our approach. 
\begin{wrapfigure}{r}{0.33\textwidth}
    \centering
    \vspace{-1.65em}
    \includegraphics[width=0.33\textwidth]{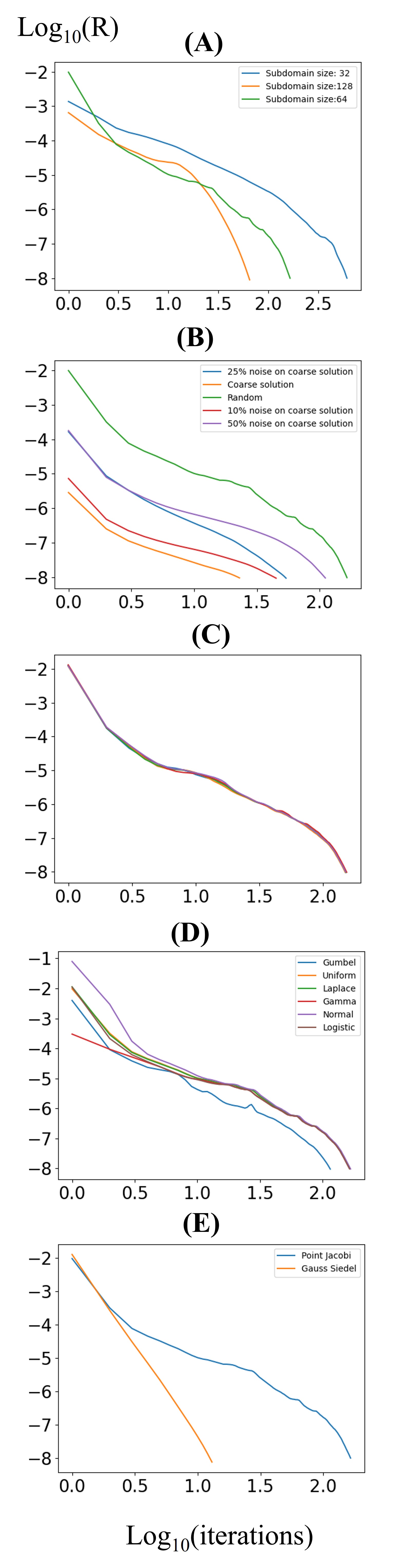}
    \vspace{-5ex}
    \setlength{\belowcaptionskip}{-78pt}
    \caption{Convergence history for (A): varying subdomain sizes; (B): coupling with traditional PDE solvers; (C, D): Robustness and stability; (E): Point Jacobi vs Gauss Seidel}
    \label{fig:10}
\end{wrapfigure}
In scenario $A$, we randomly sample $25$ initial solutions from a uniform distribution and in scenario $B$ we sample from $6$ different distributions, namely Uniform, Gumbel, Laplace, Gamma, Normal, Logistic etc. The mean convergence trajectory is plotted in Fig. \ref{fig:10}C and \ref{fig:10}D show that the log of $L_2$ norm of the convergence error falls below an acceptable tolerance for all cases and demonstrates the stability of our approach.
\subsection{CoMLSim solution methods: Point Jacobi vs Gauss Seidel}
Next, we compare the performance of the Point Jacobi and Gauss Seidel implementations in the CoMLSim approach across $3$ metrics, mean absolute error, computational cost and iterations to converge. Both methods have a similar prediction accuracy with a mean absolute error of $0.013$ and $0.017$, respectively but the Point Jacobi method is computationally cheaper requiring $4.3$ seconds to converge as compared to $72$ seconds of Gauss Seidel. It may be observed from Fig. \ref{fig:10}E that the Gauss Seidel method takes significantly fewer convergence iterations, but the Point Jacobi method is faster because it can be efficiently vectorized on a GPU.  

\subsection{Coupling with traditional PDE solvers and super-resolution}
In this experiment, we analyze the potential of coupling the CoMLSim approach with a traditional PDE solver such as Ansys Fluent and subsequently demonstrate how our approach inherently possesses the capability to perform super-resolution. 
We mainly carry out $5$ experiments, where we start from a coarse grid solution for the $100$ test samples of linear Poisson's equation solved by Ansys Fluent on a $64$x$64$ grid ($256$x smaller than the fine resolution), add random noise to the coarse initial solution with varying amplitudes ($10\%$, $25\%$ and $50\%$), and use that as an initial solution in the CoMLSim approach. The convergence trajectory in Fig. \ref{fig:10}B shows that the case with coarse grid initialization converges the fastest, followed by initialization with noise and zero initialization. All solutions converge to the same accuracy.

\subsection{Computational speed}
We observe that the CoMLSim approach is about $40$-$50$x faster as compared to commercial steady-state PDE solvers such as Ansys Fluent for the same mesh resolution and physical domain size in all the experiments presented in this work. In comparison to the ML baselines, our approach is expected to be slightly slower because it adopts an iterative inferencing approach. But, this is compensated by solution accuracy, generalizability and robustness on high-resolution grids. A detailed analysis is provided in supplementary materials section E. 

\subsection{Importance of local learning and latent-space representation}
To understand the effect of local learning, we set up the CoMLSim approach for a single subdomain (size equal to the computational domain). From this experiment, we conclude that the solution errors with a single subdomain is significantly higher than having multiple subdomains (See supplementary material section C.1.5.1). Additionally, to study the impact of latent space representations we experiment with the latent vector sizes as well as a case where CoMLSim solves in the solution space. We conclude that latent space representations have a larger impact on the computational cost than accuracy (See supplementary material section C.1.5.2). 

\subsection{Solution evolution during inference}
\begin{figure}[h]
    \centering
    \includegraphics[width=0.8\linewidth]{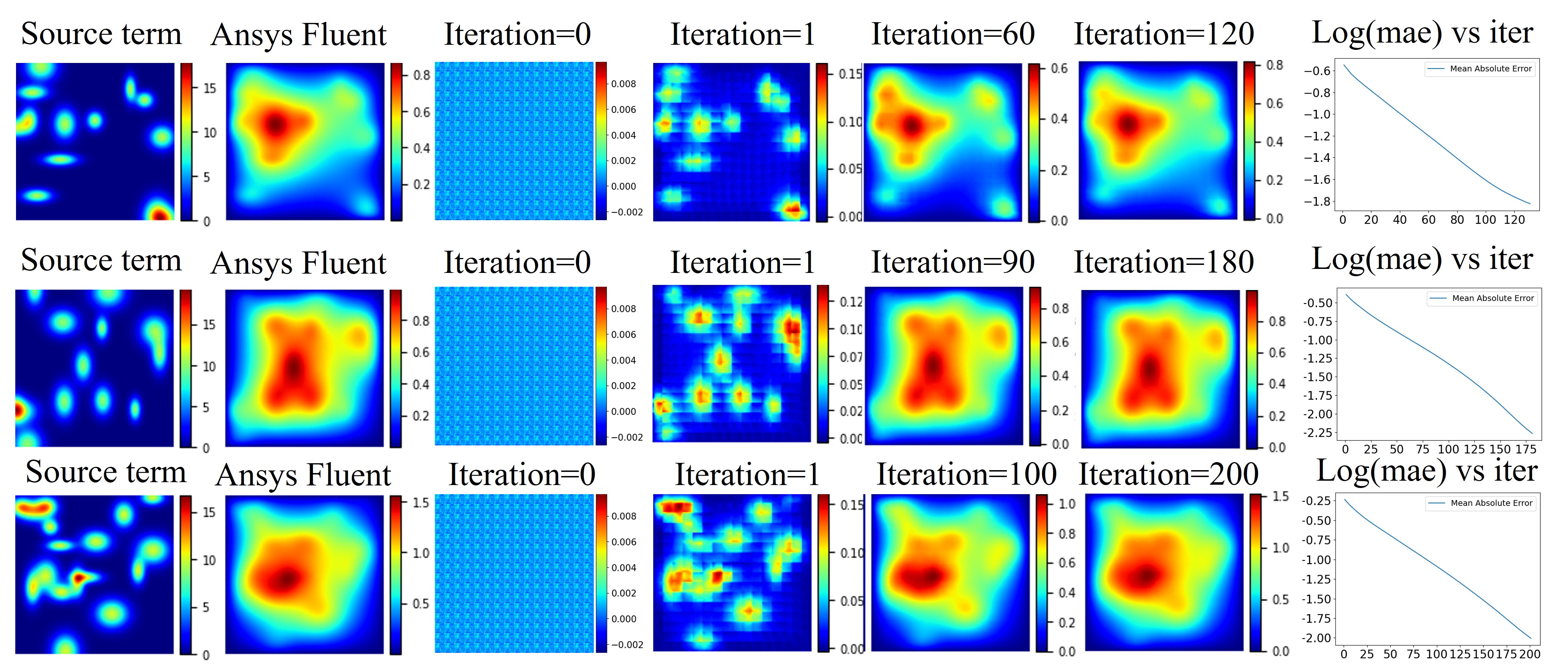}
    \caption{Evolution of PDE solution during inference}
    \label{fig:evolution}
\end{figure}

In Figure. \ref{fig:evolution}, we present the evolution of the PDE solution for the Poisson's equation experiment at different iterations of the iterative inferencing. The results are plotted for 3 unseen test cases with different source term distribution. In each case, the solution on the domain is initialized with random noise sampled from a uniform distribution between -1 and 1. The contour plot at iteration 0 shows an attenuated noisy solution because the initial solution is passed through the solution encoder and decoder. It may be observed that at iteration 1, the PDE solution starts evolving from the regions of high source term. As iterations progress, the solution begins to diffuse through the solution domain due to the repeated operation of the flux conservation network, until it converges. The diffusion process is dominant in the case of Poisson's equations and is effectively captured by the flux conservation autoencoder in CoMLSim. The error plot shows the solution accuracy at different iterations during the inference procedure. The log error is the largest initially and drops linearly as the iterations progress.
 
\section{Conclusion} \label{conclusion}
In this work, we introduce the CoMLSim approach, which is a self-supervised, low-dimensional and local machine learning approach for solving PDE on highly-resolved grids and generalizing across a wide range of PDE conditions. Our approach is inspired from strategies employed in traditional PDE solvers and adopts iterative inferencing. The proposed approach is demonstrated to predict accurate solutions for a range of PDEs, generalize reasonably across geometries, source terms and BCs. Moreover, it scales to bigger domains with larger mesh size.

\textbf{Broader impact, future work \& limitations} Although the proposed ML-model can generalize to out-of-distribution geometries, source terms and BCs, but like other ML approaches, extrapolation to any and all PDE conditions that are significantly different from the original distribution still remains a challenge. However, this work takes a big step towards laying down the framework on how truly generalizable ML-based solvers can be developed. In future, we would like to address these challenges of generalizability and scalability by training autoencoders on random, application agnostic PDE solutions and enforcing PDE-based constraints in the iterative inferencing procedure. Future work will also investigate the potential for hybrid solvers and extensions to transient PDEs and inverse problems. Finally, we will also address extension of the current approach to unstructured meshes, which is a current limitation.


%
\bibliographystyle{unsrtnat}
\bibliography{references}
\newpage

\appendix
\section*{Supplementary Materials for A composable machine-learning approach for steady-state simulations on high-resolution grids}


In the supplementary materials, we provide additional details about our approach and to support and validate the claims established in the main body of the paper. We have divided the supplementary materials into $6$ sections. Sections \ref{comlsim_arch} and \ref{baseline_arch} provide details about the network architectures and training mechanics used in the CoMLSim approach as well as the ML baselines considered in this work. This is followed by additional experimental results in Sections \ref{expt_main} and \ref{expt_additional} for PDEs considered in the main paper as well as additional canonical PDEs, namely Laplace and Darcy equations. Finally, we expand on the computational performance of CoMLSim in Section \ref{comp_perf} and provide details of reproducibility in Section \ref{reproduce}.

\section{CoMLSim network architectures and training mechanics} \label{comlsim_arch}

In this section, we will provide details about the typical network architectures used in CoMLSim followed by the training mechanics. The training portion of the CoMLSim approach corresponds to training of several autoencoders to learn the representations of PDE solutions, conditions, such as geometry and PDE source terms as well as flux conservation. In this work, we mainly employ $2$ autoencoder architectures, a CNN-based autoencoder to train the PDE solutions and conditions and a DNN-based autoencoder to train the flux conservation network.

\subsection{Solution/Condition Autoencoder}

\begin{figure}[h]
\centering
\includegraphics[width=1\linewidth]{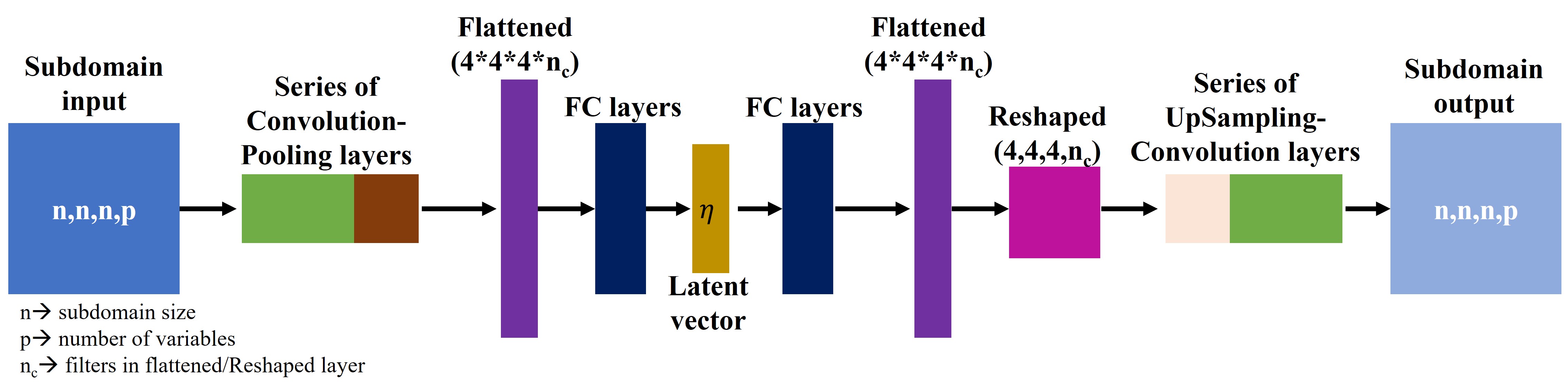}
\caption{Schematic of Autoencoder network for solution and condition}
\label{solution_encoder}
\end{figure}

These autoencoders learn to represent solutions and conditions on subdomains into corresponding lower dimensional vectors. CNN-based encoders and decoders are employed here to achieve this compression because subdomains consist of structured data representations. Figure \ref{solution_encoder} shows the architecture of a typical autoencoder used in this work to learn PDE solutions and conditions. We use separate autoencoders to learn solution and representations of conditions into lower-dimensional latent vectors. In the encoder network, we use a series of convolution and max-pooling layers to extract global features from the solution. Irrespective of the size of the input, the pooling is carried out until a resolution of $4$x$4$ in 2-D and $4$x$4$x$4$ in 3-D. This is followed by flattening and a series of dense fully-connected layers to compute the latent vector. The decoder network mirrors the encoder network exactly, except that the pooling layer is replaced by an up-sampling layer. A ReLU activation function is applied after every convolution layer. The number of filters in the convolutional layers as well as number of dense layers and the bottleneck size depends on the complexity of the application, non-linearity and sparsity in the input distribution and the size of the subdomains. 

\subsubsection{Input representation of PDE conditions}

The experiments considered in this paper have different types of PDE conditions associated with the PDE. For example, Poisson's and Non-Linear Poisson's solutions are influenced by the source term, Reynolds Averaged Navier-Stokes external flow by geometry, Darcy's solutions by diffusivity and Laplace solutions by boundary conditions. Each PDE condition is encoded into a lower-dimensional vector using the autoencoder shown in Figure \ref{solution_encoder}. Generally, diffusivity, source terms and boundary conditions have a spatial representation on the computational domain, which can be directly used to train the autoencoder. On the other hand, an efficient representation of geometry is a topic of on-going research in the ML community. Geometry can be represented using several ways such as point clouds, voxels, etc. In this work, we use the signed distance field (SDF) to represent geometry. Mathematically, the signed distance at any point within the geometry is defined as the normal distance between that point and closest boundary of a object. More specifically, for $x \in \mathrm{R}^{n}$ and object(s) $\Omega \subset \mathrm{R}^{n}$, the signed distance field $\phi(x)$ is defined by:
\[
    \phi(x)  =   
    \begin{cases} 
     + d(x, \partial \Omega) & x \in \Omega \\
     - d(x, \partial \Omega) & x \notin \Omega 
   \end{cases}  .
\]
where, $\phi(x)$ is the signed distance field for $x \in \mathrm{R}^{n}$ and objects $\Omega \subset \mathrm{R}^{n}$ \cite{gibou2018review}. \citet{maleki2021geometry} use the same representation of geometry to successfully demonstrate the encoding of geometries. However, this is a matter of choice and other valid representation can also be used in our approach.

All in all, there are two things to consider when encoding the PDE conditions, 1) the PDE condition is only encoded on subdomains that they influence and the encoding is hard-coded to a vector of zeros for all the other subdomains. For example, in an experiment of flow over a cylinder, the SDF is computed locally on each subdomain. The subdomain that cuts through the cylinder has a non-zero SDF and hence the encoding computed using the trained encoder is non-zero. Other subdomains that don't contain any parts of the cylinder can be encoded with a vector of zeros, 2) An autoencoder for PDE condition is required to be trained only if the set of conditions considered in a given problem have a spatial representation. If the PDE conditions are uniform, the magnitude can simply be considered as an encoding for a given subdomain. For example, if the source term is uniformly described on the computational domain for a given experiment, then the magnitude of the uniform source term can be used as an encoding on each subdomain. 

\subsection{Flux conservation Autoencoder}

\begin{figure}[h]
\centering
\includegraphics[width=1\linewidth]{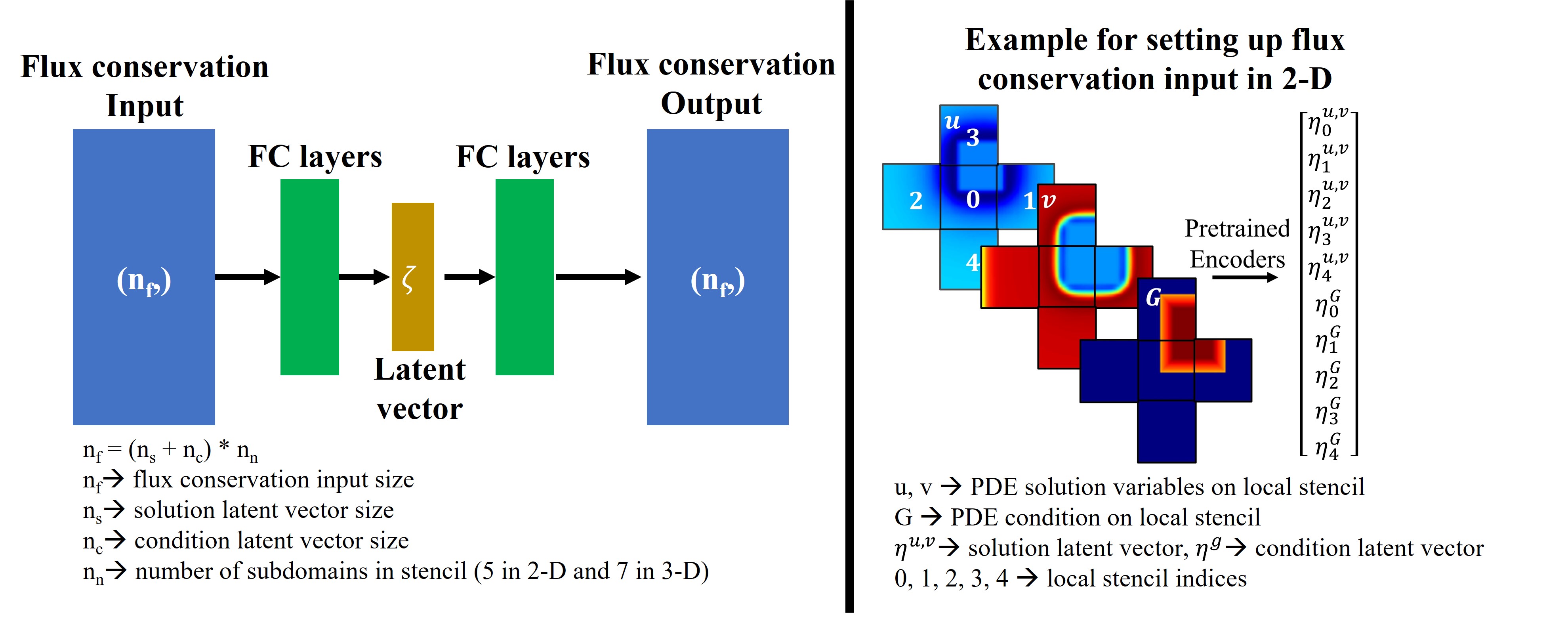}
\caption{Schematic of Autoencoder network for flux conservation}
\label{flux_encoder}
\end{figure}

These autoencoders learn to represent solution and condition encodings of a collection of neighboring subdomains. Since latent vectors don't have a spatial representation, DNN-based encoder and decoders are employed to compress them. Figure \ref{flux_encoder} shows a typical DNN-autoencoder used in this work to learn relationships between neighboring subdomains. The input to this autoencoder consist of PDE solution encoding ($\eta$) and condition encodings ($\eta_s, \eta_g, \eta_b$) on neighboring subdomains. In this work, the encoder network consists of typically $3$ hidden layers with $n, n/2, n/4$ hidden neurons, respectively. The decoder network is similar to the encoder network but the order of the hidden layers is exactly opposite. The choice of $n$, depends on the size of the input vector and the complexity of the application. In the figure, we show an example of how the input to the flux conservation network is setup for a 2-D case. Given the PDE solution and condition on a local stencil with $5$ subdomains, the encoded representations are calculated using the pre-trained encoders. The encodings of solutions and conditions are concatenated together in a pre-determined order. The same approach works in 3-D, with the difference that the local stencil has $7$ subdomains. 

\subsection{Training mechanics}
Autoencoders have overfitting tendencies and hence they are required to be trained carefully. Here, we provide general guidelines that may be used to train these autoencoders efficiently. In this work, we train all the autoencoders until an MSE of $1e^{-6}$ or an MAE of $1e^{-3}$ is achieved on a validation set. More importantly, the compression ratio is selected such that the bottleneck layer has the smallest possible size and yet satisfies the accuracy up to these tolerances. Although, each training run is very fast but may require a decent amount of hyper-parameter tuning to obtain an optimized bottleneck size. Based on the experiments and results we have shown in the main paper, the optimum performance of our approach is observed in a range of bottleneck layer sizes. But, if the bottleneck size is too small or too large, the performance deteriorates. All the autoencoders are trained with the NVIDIA Tesla V-100 GPU using TensorFlow. The autoencoder training is a one-time cost and is reasonably fast.

\color{black}
\subsection{Similarities between CoMLSim and Traditional PDE solvers}

In this section we expand on the main similarities in between our approach on a traditional Finite Volume or Finite Difference based PDE solvers. There are 3 main similarities, 1) Domain discretization, 2) Flux conservation and 3) Iterative solution algorithm. Here we provide more details about each one.

\begin{wrapfigure}{r}{0.23\textwidth}
        \centering
        \vspace{-2em}
        \includegraphics[width=3cm]{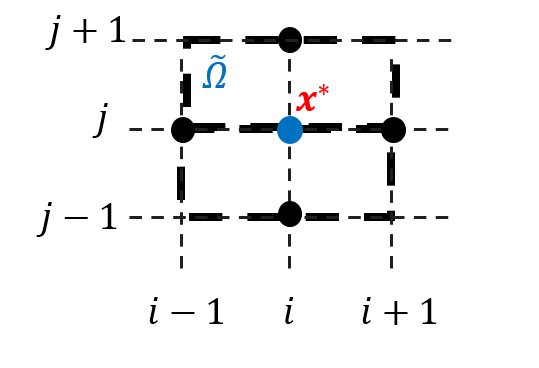}
        \vspace{-2ex}
        \caption{2D stencil}
        \label{fig:stencil}
\end{wrapfigure}

Consider a set of coupled PDEs with $n$ solution variables. For the sake of notation simplicity, we take $n=2$, such that $u(x,y,z,t)$ and $v(x,y,z,t)$ are defined on a computational domain $\Omega$ with boundary conditions specified on the boundary of the computational domain, $\Omega_b$. It should be noted that extension to more solution variables is trivial.
The coupled PDEs are defined as follows:
\begin{equation}
        L_1(u, v) - F_1  = 0; L_2(u, v) - F_2 = 0
    \label{eq1u}
\end{equation}

where, $L_1,$ $L_2$ denote PDE operators and $F_1,$ $F_2$ represent PDE source terms. The PDE operators can vary for different PDEs. For example, in a non-linear PDE such as the unsteady, incompressible Navier-Stokes equation the operator, $L = \frac{\partial}{\partial t} + \vec{a}.\vec{\nabla} - \vec{\nabla}.\vec{\nabla}$

\begin{enumerate}
    \item \textbf{Domain discretization:} Traditional PDE solvers solve PDEs given in Eq. \ref{eq1u} by representing solutions variables, u, v, and their spatio-temporal derivatives on a discretized computational domain. The domain is discretized into a finite number of computational elements, using techniques such as Finite Difference Method (FDM), Finite Volume Method (FVM) and Finite Element Method (FEM). 

    Similar to traditional PDE solvers, the first step in the CoMLSim is to decompose the computational domain into smaller subdomains. A single subdomain in the CoMLSim is analogous to a computational element in the traditional solver because the CoMLSim predicts PDE solutions directly on local subdomains.
    
    \item \textbf{Flux conservation:} Traditional PDE solvers use numerical approximation schemes are used to compute linear and non-linear components of the PDE. For example, in Eq. \ref{eq1u}, if $L = \frac{\partial u}{\partial x} + \frac{\partial v}{\partial y}$, representing the 2-D incompressible continuity equation in fluid flows, the spatial derivatives can be approximated on a uniform stencil shown in Figure \ref{fig:stencil} using a second order Euler approximation shown below in Eq. \ref{eq2u}.
    
    \begin{equation}
            L_{i,j} = \frac{u_{i+1,j}-u_{i-1,j}}{2\Delta x} + \frac{v_{i, j+1}-v_{i,j-1}}{2\Delta y}
        \label{eq2u}
    \end{equation}
    where $i, i+1, i-1, j, j+1, j-1$ are element indices and $\Delta x, \Delta y$ correspond to the size of stencil. These numerical approximations denote flux conservation between neighboring elements. Fluxes represent the flow of information between neighbors and hence, their accurate representation is crucial for information propagation within the domain.
    
    Similarly, flux conservation in the CoMLSim happens across neighboring subdomains to ensure local consistency and information propagation. The representation of PDE discretization on subdomains is similar to Equation Eq. \ref{eq2u} but the indices i, j represent subdomain indices and the numerical schemes for discretization are now represented by a neural network, Theta, as shown in Eq. \ref{eq3u}.
    
    \begin{equation}
        L_{i,j} = \Theta(\eta^u_{i,j}, \eta^u_{i+1,j},\eta^u_{i-1,j},\eta^u_{i,j+1},\eta^u_{i,j-1}, \eta^v_{i,j}, \eta^v_{i+1,j},\eta^v_{i-1,j},\eta^v_{i,j+1},\eta^v_{i,j-1})
        \label{eq3u}
    \end{equation}
    where, $\eta_u$, $\eta_v$ are encodings of solution fields on subdomains and $i, i+1, i-1, j, j+1, j-1$ are subdomain indices.
    
    \item \textbf{Iterative solution procedure:} In traditional solvers, the discretized PDEs represent a system of linear or non-linear equations, where the number of such equations equals the number of computational elements. To solve the PDE solutions, the discretized PDE residual is minimized by enforcing flux conservation iteratively using linear and non-linear equation solvers.

    Similar to traditional solvers, the discretized PDEs represent a system of linear or non-linear equations, where the number of such equations equals the number of computational subdomains. To solve this system of equations we employ exactly the same techniques that a traditional would use. For example, in this work we have explored 2 linear iterative solution methods, such as Point Jacobi and Gauss Seidel. 

\end{enumerate}

\subsection{Self-supervised solution algorithm}

Although, this approach requires solution samples to train the autoencoders, we claim in the paper that it is self-supervised in the sense that we don’t use these samples to learn an explicit relationship between the input and output distribution. Our training consists of simply training autoencoders and the inference algorithm involves solving a constrained fixed-point iteration to converge to a PDE solution. In the constrained fixed-point iteration, the solution converges to a PDE solution starting from initial random noise. Our solution algorithm is never taught this trajectory of solution convergence but discovers that by itself. Hence, we claim that the solution algorithm at inference is self-supervised.

\color{black}

\section{Description of baseline network architectures} \label{baseline_arch}

In the main paper, we compare the performance of CoMLSim with UNet \cite{ronneberger2015u}, FNO \cite{li2020fourier}, DeepONet \cite{lu2021learning} and FCNN \cite{zhu2018bayesian}. Here we describe the network architectures used to train the respective models for all the experiments considered in the main paper.

\textbf{UNet \cite{ronneberger2015u}:} The encoder part of the network has $10$ convolutional blocks, $2$ at each down-sampled size. The input is down-sampled by $4$x. The decoder part of the network predicts the output by up-sampling the bottleneck and using skip connections from the encoder network by concatenating the corresponding upsampled output with the corresponding down-sampled output. The decoder part of the network also has $10$ convolutional blocks, $2$ after each up-sampled size and has $256$ stacked channels. The total number of learnable parameters in UNet baseline is equal to 0.471 million in 2-D and 1.412 million in 3-D.

\textbf{Fourier Neural Operator (FNO) \cite{li2020fourier}:} The FNO model is same as the original implementation in \citet{li2020fourier} but the number of modes are increased to $8$ for 3-D experiments to achieve better training loss. The FNO model has 1.188 million parameters in 2-D and 3.689 million parameters in 3-D. 

\textbf{DeepONet \cite{lu2021learning}:} The DeepONet architecture has two branches, a branch net and a trunk net. In all cases, the trunk net has $3$ hidden layers with $512$ neurons each. The branch is a convolutional neural network, which takes inputs the spatial source term. It is extremely difficult to train the DeepONet with the full resolution of the PDE conditions because of the massive data storage requirements. Hence, for all experiments the PDE conditions are uniformly subsampled to a lower grid resolution given as an input to the branch net. The branch net has $4$ DownSample blocks and $10$ convolutional blocks, $2$ at each down-sampled size. Additionally, the DeepONet is extremely sensitive to the sampling strategy adopted in the training data. The total number of learnable parameters is equal to 1.353 million. The subsampling of PDE conditions and the random sampling used in this work may have affected the testing accuracy of DeepONet. 

\textbf{Fully Convolutional Neural Network (FCNN) \cite{zhu2018bayesian}:} The FCNN model is similar to the original implementation in \citet{zhu2018bayesian} but the number of convolution filters and downsampling layers are tuned to accommodate the high-resolutions and non-linearity in different use cases. The FCNN model has 0.189 million parameters in 2-D and 0.578 million parameters in 3-D. 

\section{Experiments results and details from main paper} \label{expt_main}
We demonstrated the CoMLSim for $4$ experiments in the main paper. Here we provide more details about the CoMLSim setup as well as additional results and discussions for each experiment. \color{black}The different experiments presented in this work are a good mix of pure research and engineering problems with varying levels of non-linear complexity, input distributions, PDEs, solution variable coupling, spatial dimension etc. It must be noted that CoMLSim as well as the baselines are trained with reasonably training samples. The baseline performance can have different outcomes with increasing the size of the training data.
\color{black}
\subsection{2-D Linear Poisson's equation} \label{linear_poisson}

The Poisson's equation is shown below in Eq. \ref{eqc1}.

\begin{equation}
        \nabla^2 u = f
    \label{eqc1}
\end{equation}

where, $u$ is the solution variable and $f$ is the source term. In this experiment, the source term is sampled from a Gaussian mixture model, where the number of Gaussians is randomly chosen between $1$ and $30$ and each Gaussian has a randomly specified mean and standard deviation. The Gaussian mixture model is described below in Eq. \ref{eqc1}. The computational domain is 2D and is discretized with a highly-resolved grid of resolution $1024$x$1024$. The high-resolution grid is required in this case to capture the local effects of the source term distribution.

\begin{equation}
        \sum_{j=0}^{1024}\sum_{j=0}^{1024} f_{i,j} = \sum_{j=0}^{1024}\sum_{j=0}^{1024}\sum_{k=0}^{30} A_k\exp\left( -\frac{x-\mu_{x,k}}{\sigma_{x,k}} - \frac{y-\mu_{y,k}}{\sigma_{y,k}} \right)
    \label{eqc2}
\end{equation}
where, $x, y$ correspond to the grid coordinates. $A_k$ randomly assumes either $0$ or $1$ to vary the number of active Gaussians in the model. $\mu_x, \mu_y$ and $\sigma_x, \sigma_y$ are the mean and standard deviations of Gaussians in $x$ and $y$ directions, respectively. The means vary randomly between $0$ and $1$, while the standard deviations are varied between $0.001$ and $0.01$. The smaller magnitude of standard deviation results in hot spots that can only be captured on highly-resolved grids. 

\subsubsection{Training} $256$ solutions are generated for random Gaussian mixtures using Ansys Fluent and used to train the different components of CoMLSim. The computational domain of $1024$x$1024$ resolution is divided into $256$ subdomains each of resolution $64$x$64$. The solution and source terms are compressed into latent vectors of size $11$, respectively. The flux conservation autoencoder has a bottleneck layer of size $35$.

\subsubsection{Testing} $100$ more solutions are generated for random Gaussian mixtures using Ansys Fluent. Due to the high-dimensionality of the source term description, the testing set has no overlap with the training set. The convergence tolerance of the CoMLSim solution algorithm is set to $1e^{-8}$ and the solution method is Point Jacobi. Each solution converges in about $2$ seconds \color{black}and requires about 150 iterations on an average.\color{black}

\subsubsection{Comparisons with Ansys Fluent for in-distribution testing}
\begin{figure}[h!]
\centering
\includegraphics[width=0.5\linewidth]{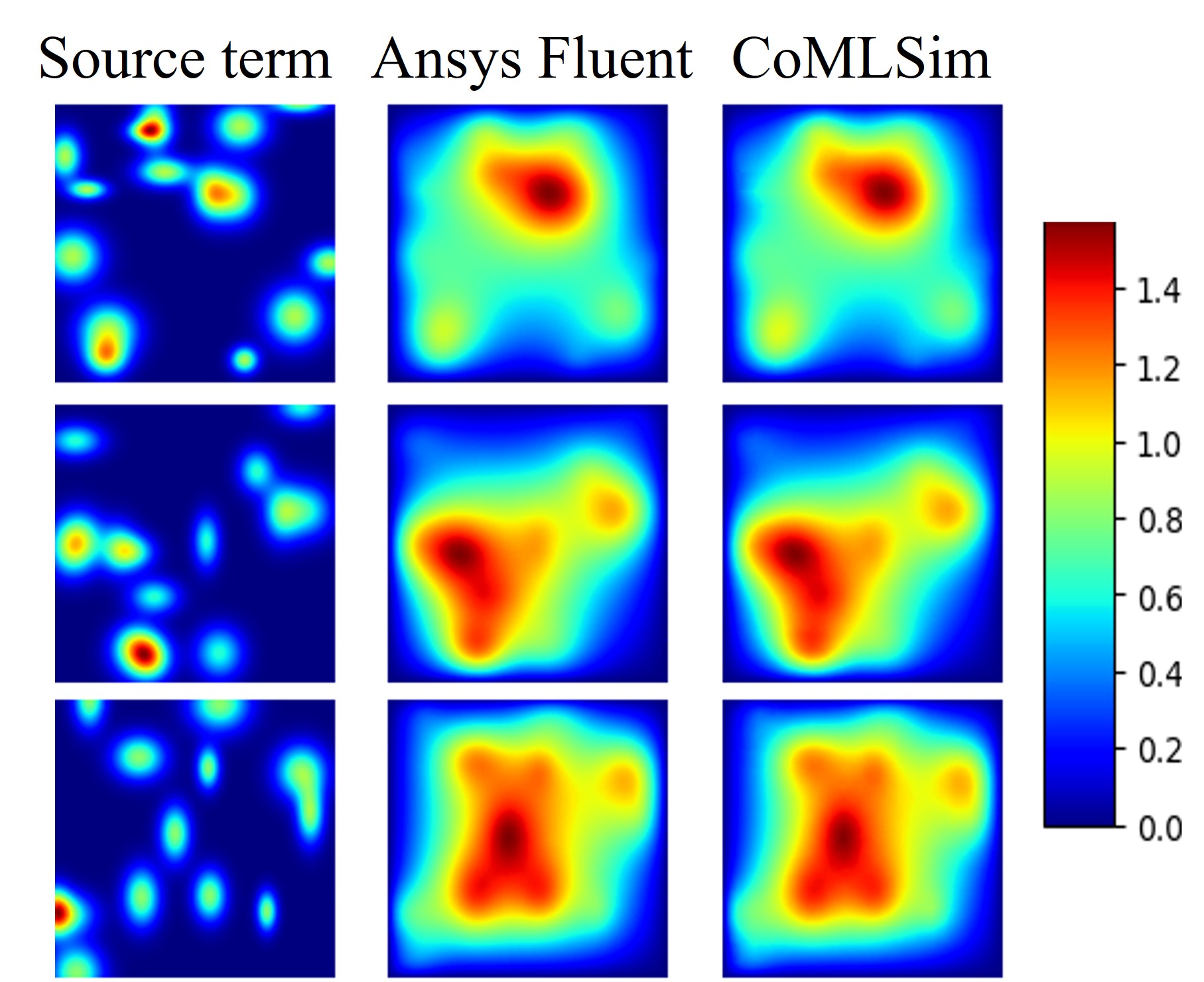}
\caption{CoMLSim vs Ansys Fluent for linear Poisson's equation with unseen in-distribution source terms}
\label{fig:c1}
\end{figure}
The CoMLSim predictions for a selected unseen testing samples are compared with Ansys Fluent in Fig. \ref{fig:c1}. It may be observed that the contour comparisons agree well with Fluent solutions. Overall, the mean absolute error over $100$ testing samples is $0.011$. 

\subsubsection{Comparisons with Ansys Fluent for out-distribution testing}

In this section, we demonstrate the generalization capability of CoMLSim as compared to Ansys Fluent and all the baselines. We have not presented results from DeepONet because it is does not perform as well as the other baselines. We consider $3$ experiments outlined below.

\begin{enumerate}
    \item \textbf{Higher number of Gaussians:}
    It may be observed from Eq. \ref{eqc2} that the maximum number of Gaussians allowed is set to $30$ and this number is sampled from a uniform distribution. In this experiment, we fix the number of Gaussians to $30$, $40$, $50$ and $60$ and generate $10$ solutions for each case using Ansys Fluent. As the number of Gaussians increase the source term distribution moves further away from the distribution in Eq. \ref{eqc2} used in training. 
    \begin{figure}[h]
    \centering
    \includegraphics[width=0.8\linewidth]{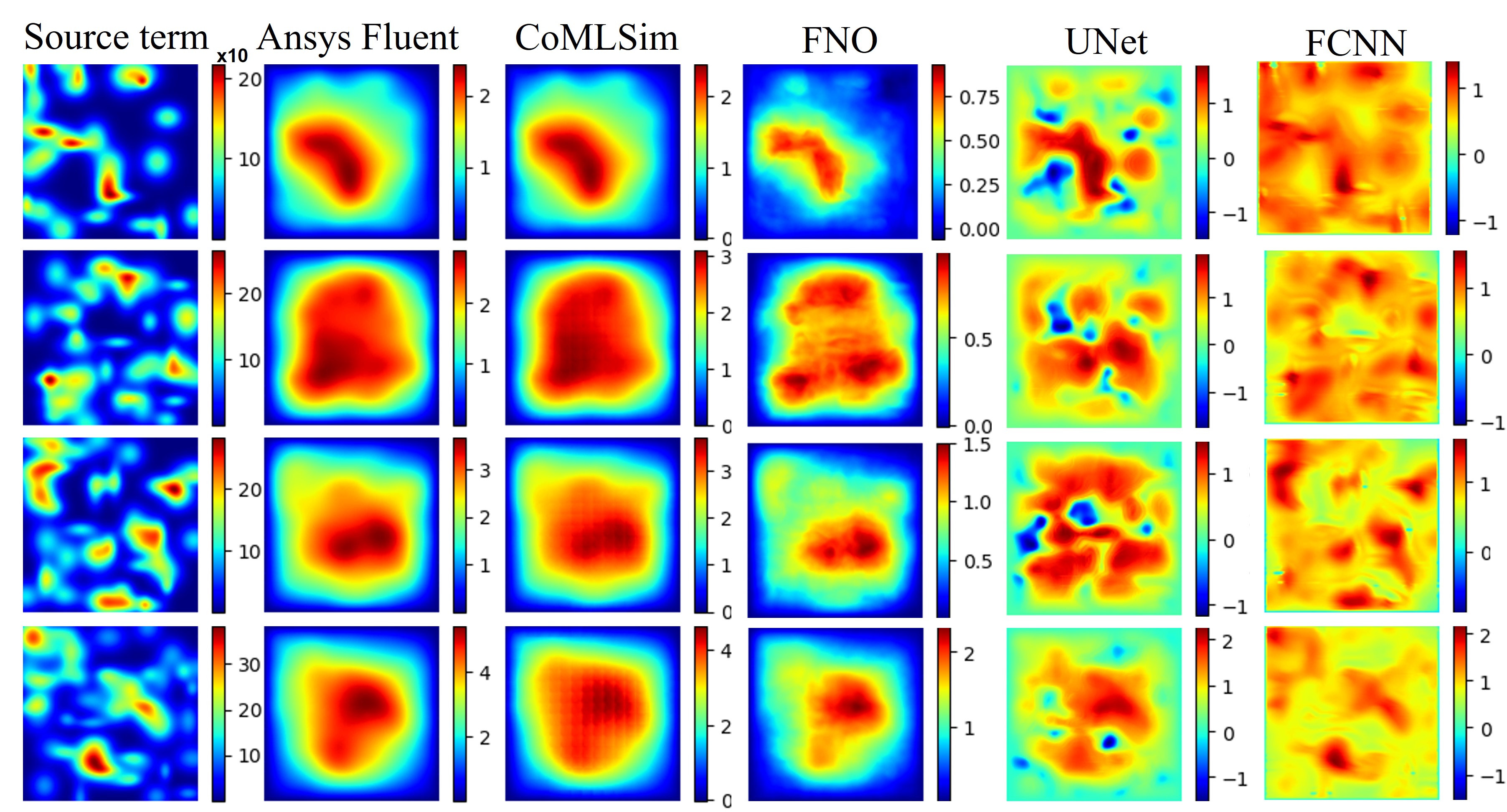}
    \setlength{\belowcaptionskip}{-8pt}
    \caption{Comparison for linear Poisson's equation on out-of-distribution source terms. Number of Gaussians vary from $30$ to $60$ from top to bottom.}
    \label{fig:c2}
    \end{figure}
    
    It may be observed from the contour plots in Figure \ref{fig:c2}, that the CoMLSim approach consistently beats all the ML baselines. As the number of Gaussians increases, the total source term applied on the computational domain increases proportionally. As a result, the magnitude of the solution is also higher and this can be observed in the color map scale in Figure \ref{fig:c2}. The CoMLSim captures the both the spatial solution pattern as well as the magnitude correctly in comparison to all ML-baselines. Amongst the ML baselines, FNO performs the best.
    
    \item \textbf{Completely different source distribution:}
    In this experiment, we sample the source term from a completely different distribution. The source term distribution is discontinuous, where the computational domain is divided into either $8^2$ or $16^2$ tiles and a uniform source term is specified on each tile such that the total source is conserved. The resulting discontinuity in source terms between neighboring tiles makes it challenging to calculate the solution, even with traditional PDE solvers. We generate $10$ solutions each with $8^2$ and $16^2$ tiles using Ansys Fluent. This source term distribution is significantly different from the Gaussian distribution that was initially considered for training. 
     
    \begin{figure}[h]
    \centering
    \includegraphics[width=0.8\linewidth]{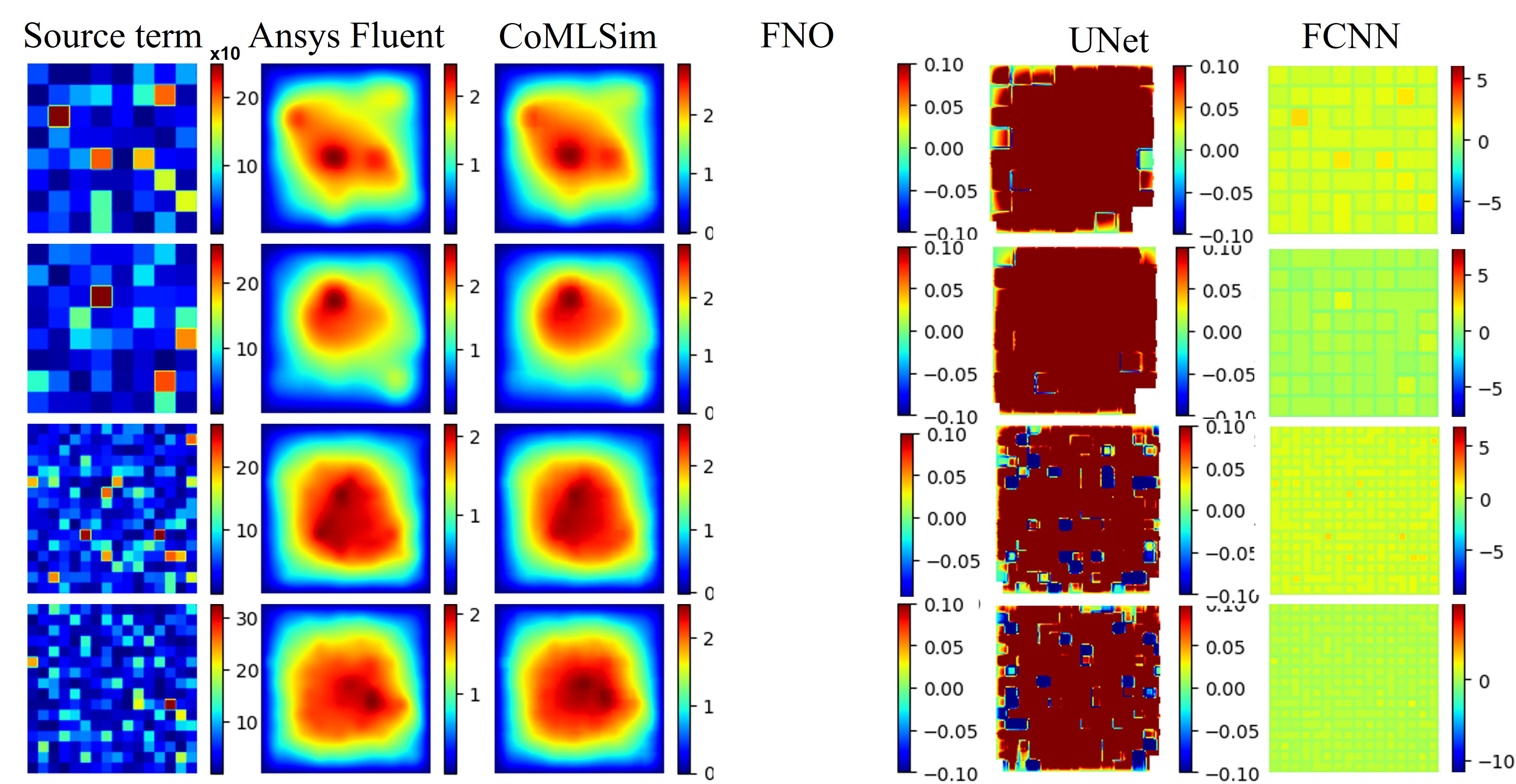}
    \caption{Comparison for linear Poisson's equation on out-of-distribution discontinuous source terms}
    \label{fig:c3}
    \end{figure}
    
    It may be observed from the contour plots in Figure \ref{fig:c3}, that the ML baselines don't predict a reasonable solution. On the other hand, the CoMLSim predicts solutions with higher accuracy. This can be attributed to the local and latent space learning strategies adopted which enables it to accurately learn and predict the local physics of the PDE. It may also be observed that FNO predicts NaN values because the Fourier transform of the discontinuous source term in the first layer of FNO results in NaN.

    \item \textbf{Bigger domain with mesh size of 2048x2048:}
    
    In the main paper, we showed how CoMLSim is designed to scale to bigger physical domains with larger mesh sizes. Here, we present comparisons at a mesh size of $2048^2$ between CoMLSim, Ansys Fluent and the ML baselines for the first $4$ samples in the testing set. 
    \begin{figure}[h]
    \centering
    \includegraphics[width=0.8\linewidth]{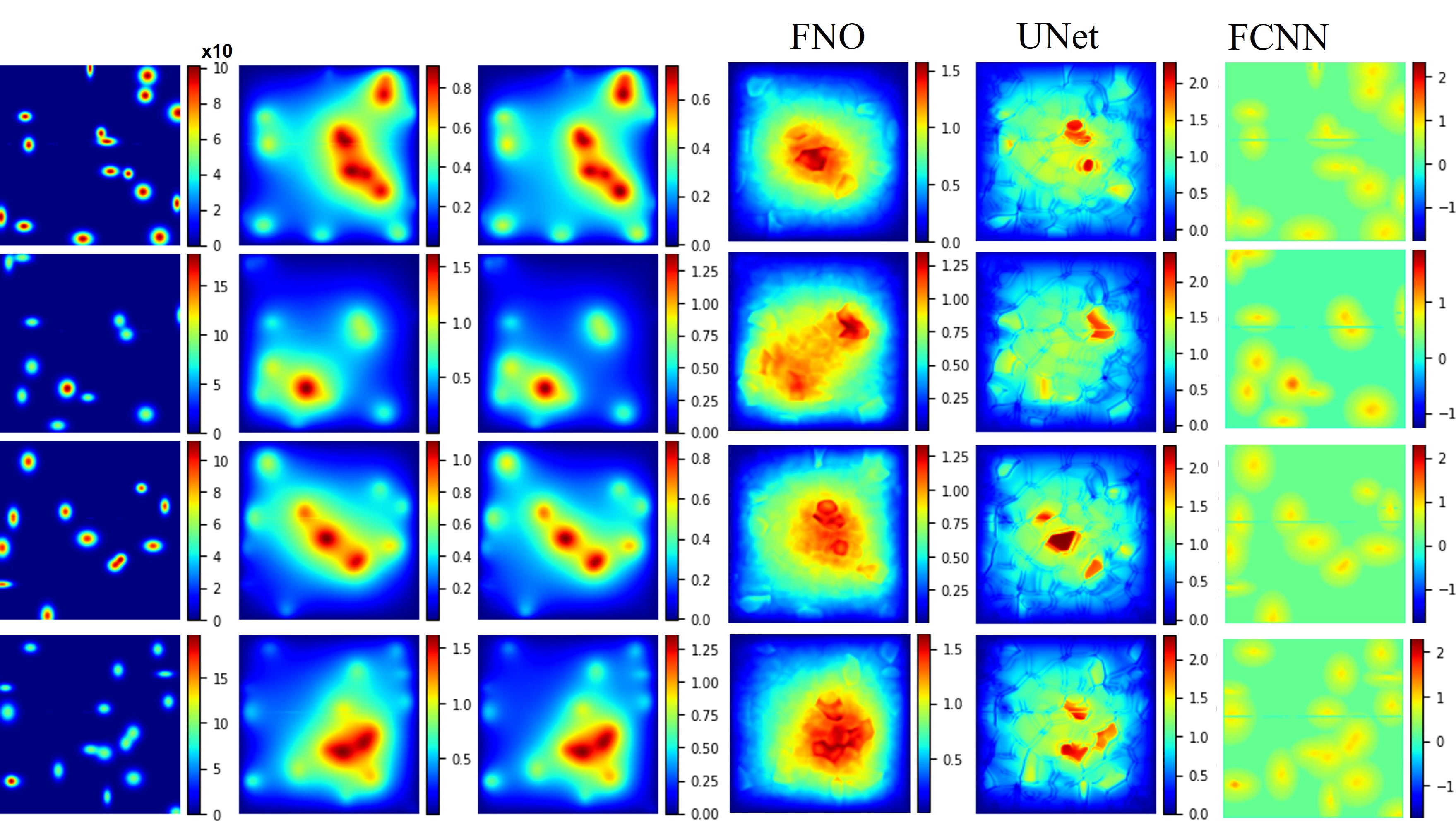}
    \setlength{\belowcaptionskip}{-8pt}
    \caption{Comparison for linear Poisson's equation for a bigger domain with mesh size of 2048x2048}
    \label{fig:c4}
    \end{figure}
    It may be observed from Figure \ref{fig:c4} that the CoMLSim outperforms all baselines and matches well with Ansys Fluent.
\end{enumerate}

\color{black}
\subsubsection{Additional Ablation Studies}

In this section we carry out additional ablation studies to further understand the different aspects of the CoMLSim approach. The CoMLSim has $3$ main features, 1) Local learning, 2) Latent space representations and 3) iterative inferencing. The ML components in our approach, such as the solution, condition and flux conservation autoencoders are designed to support these features. In this section, we present experiments to understand the impact of each of these features on the performance of our approach.

\textbf{1) Importance of local learning:} 

In this experiment we setup the CoMLSim with a single subdomain. The solution and source term autoencoders, in this case, are trained on the entire computational domain (1024x1024). Since there are no neighbors, the flux conservation autoencoder is trained on a 1-subdomain stencil. The CoMLSim approach is tested on 250 samples. The mean absolute error obtained is equal to 0.072. The cases reported in the paper in Section 4.3 with 32, 64 and 128 subdomains have mean absolute errors of 0.015, 0.011 and 0.029, respectively. In comparison to Table 1 from the paper, it may be observed that the errors from the single subdomain CoMLSim approach are similar to the ML-baselines. On the other hand, the computational time required by the single subdomain instance is about 3x more expensive than multi-subdomain instance of CoMLSim. This is due to large latent vector sizes obtained from encoding solutions on highly resolved computational domains (1024x1024) as opposed to using subdomains (64x64). Moreover, single subdomain CoMLSim instance cannot scale to bigger computational domains with larger meshes.

\textbf{2) Importance of latent space representation:} 

In this experiment, we set the encoder and decoder of solution and condition autoencoders to identity. By doing this we show that as we reduce to a standard domain decomposition, the accuracy of the approach is maintained but the computational time significantly increases. As stated earlier, we use a subdomain resolution of 64x64 to discretize the domain of 1024x1024 resolution into 256 smaller subdomains. With identity encoders and decoders, the latent size of solutions and source terms on each subdomain is equal to 4096, which is the number of computational elements in a subdomain. As a result, the flux conservation autoencoder has an input of (4096 + 4096) * 5 = 40960, where 5 is the number of surrounding neighbors. Training a fully connected neural network with 40960 input features requires networks with billions of parameters and computationally prohibitive GPU memory. Due to these reasons, we reduce the subdomain size to 16x16 to generate a comparison for accuracy. The flux conservation compression ratio (bottleneck size / input size) is consistent with the experiment in the paper.  The accuracy observed on 250 testing samples is around 0.017 and is consistent with the results reported Table 1. However, the computational time required in this case is about 2-3x larger than the results reported in Section \ref{comp_perf}.

\textbf{3) Impact of latent vector size:} 

In this experiment we evaluate the impact of solution and source term latent sizes with accuracy and computational time of CoMLSim approach averaging over 250 testing samples. It may be observed from the table \ref{tab:solution_latent} that the accuracy decreases as we increase the latent sizes. This is due to overfitting of the autoencoder, which makes the latent vector less pronounced. The computational iterations also increase. This observation is consistent with the experiment in Section 4.4 of the main paper where we vary the size of the latent size in flux conservation autoencoder. 

It is important to note that the result of this experiment cannot be compared to standard domain decomposition. The solution and condition autoencoders in the case of standard domain decomposition were identity and resemble error-free encoding. In this case, autoencoders are overfit for larger latent sizes and substantially diminish the prominence of the latent vector, thereby resulting in bad performance.

\begin{table}[h]
\caption{Solution/Condition Encoding size vs accuracy and number of convergence iterations}
\centering
\begin{tabular}{ |p{4cm}|p{4cm}|p{4cm}|  }
\hline
Encoding size& Mean Absolute Error & Avg. Iterations\\
\hline
7 & 0.047 & 75\\
11   & 0.017   & 150 \\
32 &   0.029 & 250 \\
64 &   0.1 & 600 \\
128 &   0.21 & 700 \\
256 &   0.29 & 800 \\
\hline
\end{tabular}
\label{tab:solution_latent}
\end{table} 


 

\color{black}
\subsection{2-D Non-linear coupled Poisson's equation} \label{nonlinear_poisson}

The coupled non-linear Poisson's equation is shown below in Eq. \ref{eqc3}.
\begin{equation}\label{eqc3}
\left. \begin{array}{ll}  
\quad\quad\quad\quad \quad\quad\displaystyle\nabla^2 u = f - u^2\quad\quad \quad\quad\quad\quad\quad\quad\quad\quad\\
 \quad\quad\quad\quad \quad\displaystyle\nabla^2 v = \frac{1}{u^2+\epsilon} - v^2\quad\quad\quad\quad\\[8pt]
 \end{array}\right\}
\end{equation}
where, $u, v$ are the solution variables and $f$ is the source term. The source term is similar to the experiment in Section \ref{linear_poisson} except that the $x, y$ standard deviations are varied between $0.001$ and $0.05$. As opposed to linear Poisson's, in this problem the complexities result from the coupling of solution variables as well as the source term distribution. A closer look at the PDE in Eq. \ref{eqc3} shows us that the variable $v$ is implicitly coupled with the source term and this relationship is challenging to discover for ML methods. 

\subsubsection{Training} $256$ solutions are generated for random Gaussian mixtures using Ansys Fluent and used to train the different components of CoMLSim. The computational domain is divided into $256$ subdomains each of resolution $64$x$64$. The solution and source terms are compressed into latent vectors of size $11$, respectively. The flux conservation autoencoder has a bottleneck layer of size $35$.

\subsubsection{Testing} $100$ more solutions are generated for random Gaussian mixtures using Ansys Fluent. The convergence tolerance of the CoMLSim solution algorithm is set to $1e^{-8}$ and the solution method is Point Jacobi. Each solution converges in about $2$ seconds \color{black}and requires about 150 iterations on an average.\color{black}

\subsubsection{Comparisons with Ansys Fluent for in-distribution testing}
\begin{figure}[h]
\centering
\includegraphics[width=0.6\linewidth]{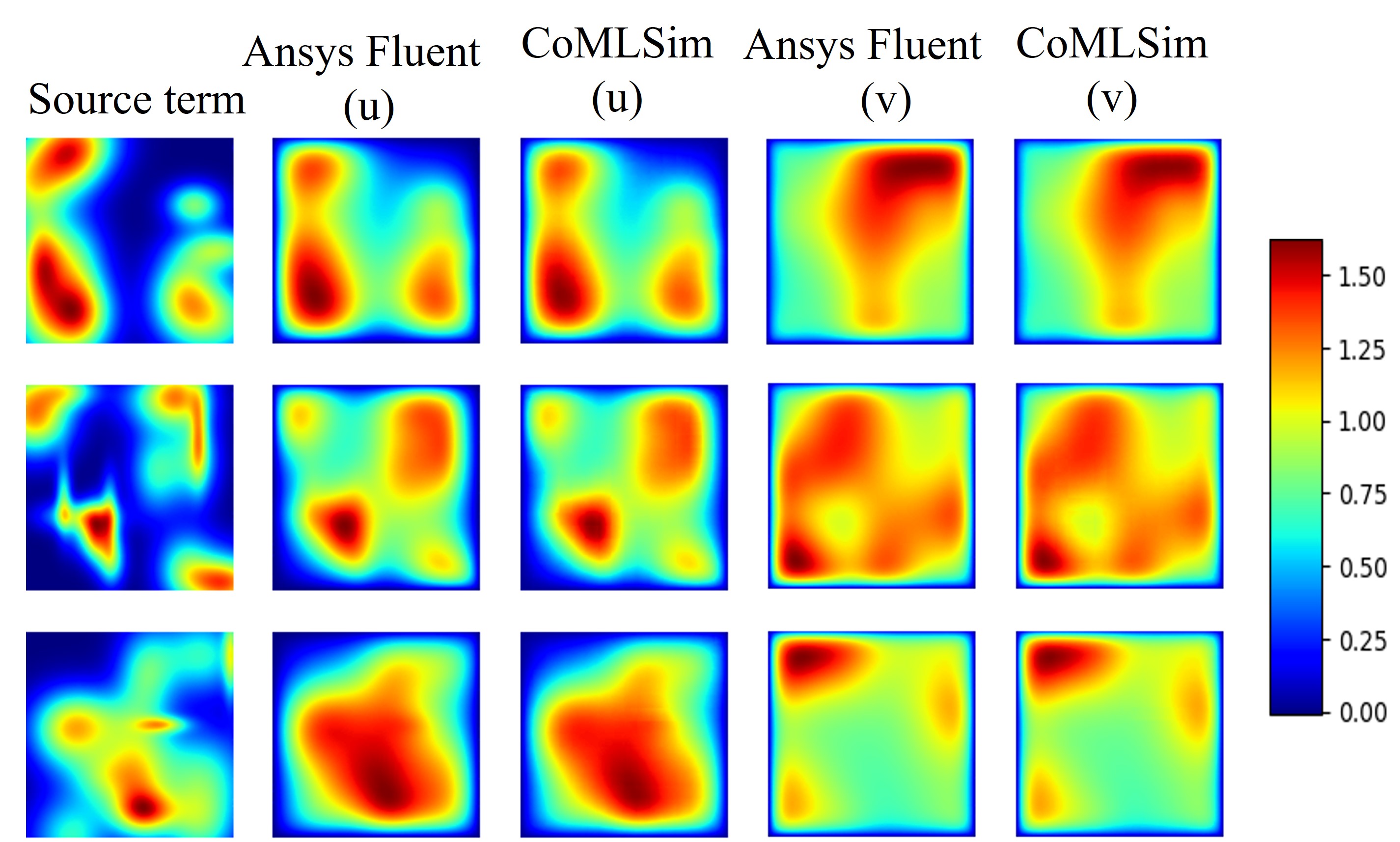}
\setlength{\belowcaptionskip}{-8pt}
\caption{CoMLSim vs Ansys Fluent for non-linear Poisson's equation for in-distribution unseen source terms}
\label{fig:c4}
\end{figure}

The CoMLSim predictions for a selected unseen testing samples are compared with Ansys Fluent in Fig. \ref{fig:c4}. It may be observed that the contour comparisons agree well with Fluent solutions. Overall, the mean absolute error over $100$ testing samples is $0.0053$. The mean absolute error is smaller than the linear Poisson experiment because of the smoothness introduced in the power map by increasing the Gaussian mixture standard deviation. Comparisons with other ML baselines are provided in the main paper. 

\subsubsection{Comparisons with Ansys Fluent for out-distribution testing}

In this section, we demonstrate the generalization capability of CoMLSim as compared to Ansys Fluent and all the baselines. Similar to the generalizability test with linear Poisson's, we fix the number of Gaussians to $30$, $40$, $50$ and $60$ and generate $10$ solutions for each case using Ansys Fluent. As the number of Gaussians increase the source term distribution moves further away from the distribution in Eq. \ref{eqc2} used in training. 

\begin{figure}[h!]
\centering
\includegraphics[width=0.7\linewidth]{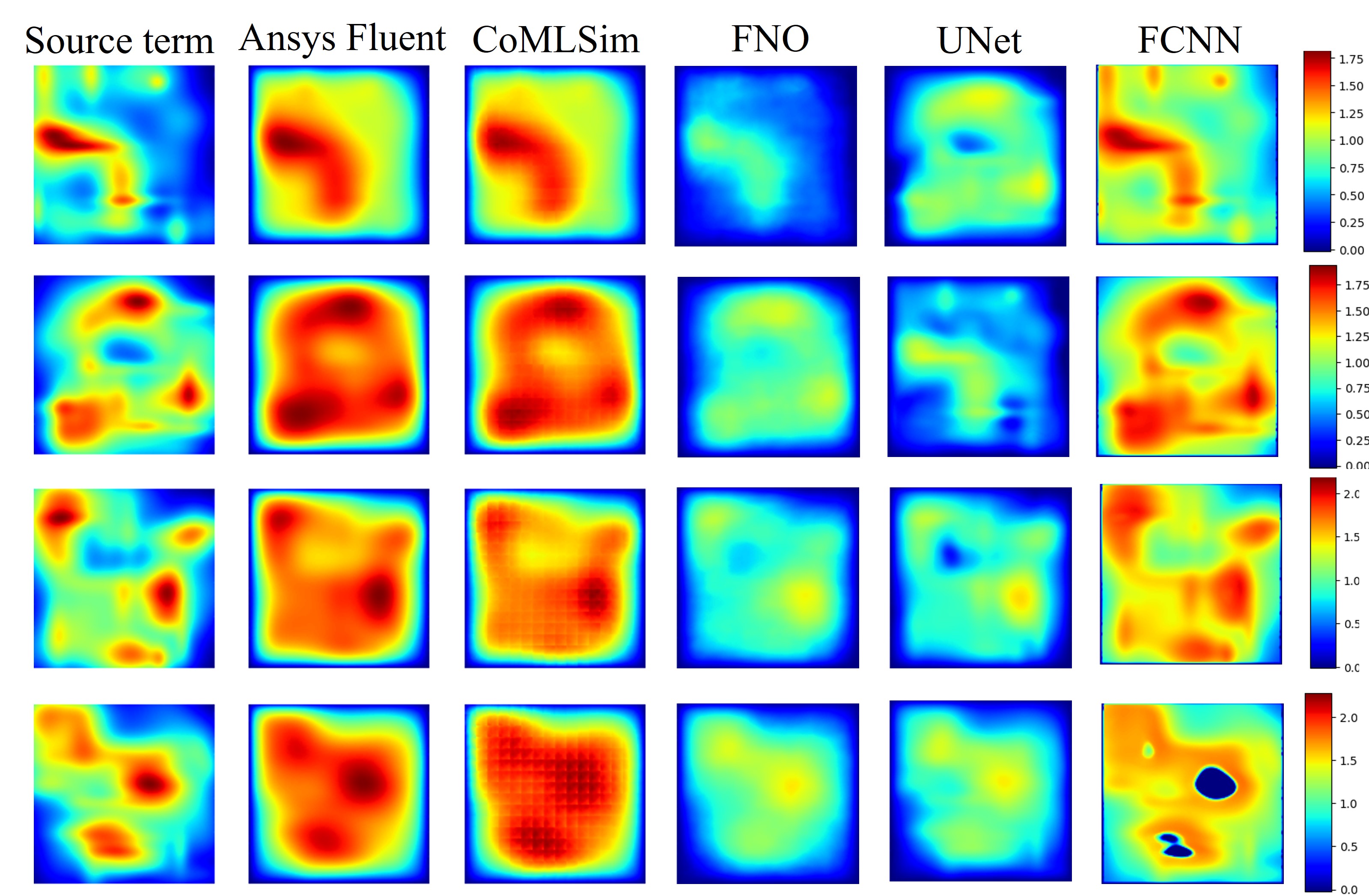}
\setlength{\belowcaptionskip}{-8pt}
\caption{Comparison for variable $u$ of non-linear Poisson's equation for out-of-distribution source terms. The number of Gaussians vary from 30 to 60 from top to bottom.}
\label{fig:c5}
\end{figure}
\begin{figure}[h!]
\centering
\includegraphics[width=0.7\linewidth]{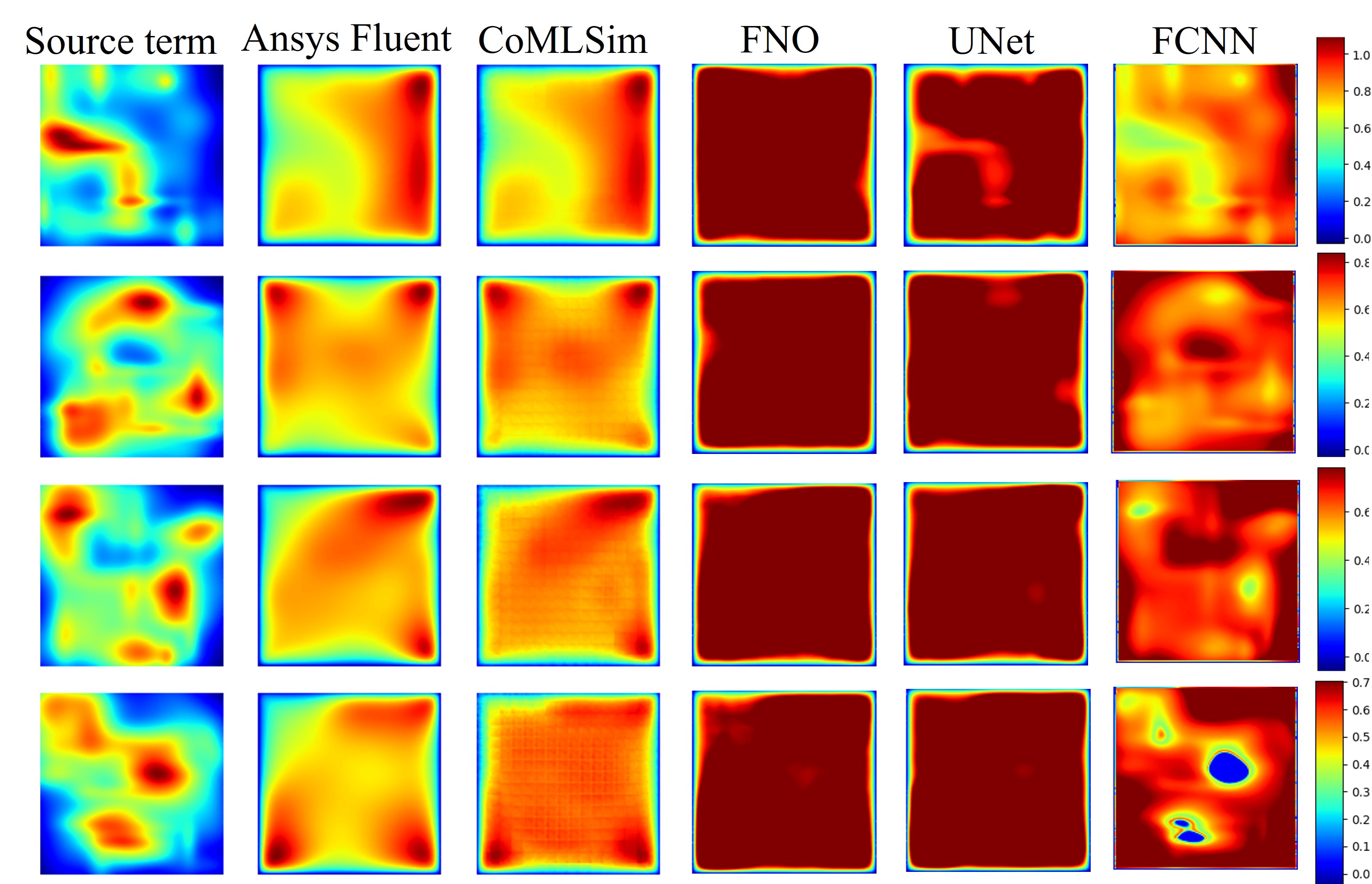}
\setlength{\belowcaptionskip}{-8pt}
\caption{Comparison for variable $v$ of non-linear Poisson's equation for out-of-distribution source terms. The number of Gaussians vary from 30 to 60 from top to bottom.}
\label{fig:c6}
\end{figure}
It may be observed from Figures \ref{fig:c5} and \ref{fig:c6} that the CoMLSim outperforms all the ML baselines and matches well with Ansys Fluent for both solution variables. The out-of-distribution source term has a much stronger impact on the coupled solution variable as compared to linear Poisson's experiment. It may be observed from the figures that the ML-baselines predict reasonably for solution variable $u$, but the errors are much higher for the coupled variable, $v$. 

\subsection{3-D Reynolds Averaged Navier-Stokes external flow} \label{ns_ext}

In this experiment, we apply the CoMLSim approach to model high Reynolds number turbulent flow around arbitrary geometry shapes built for primitive objects such as Cylinder, Cuboid, NURBS etc and their combinations. The governing PDEs for this case are shown in Eq.\ref{eqc6} 

\begin{equation}\label{eqc6}
\left. \begin{array}{ll}  
\mbox{\textbf{Continuity:} } \quad\quad\quad\quad\quad\quad\quad\quad\quad\quad\quad \quad\quad\displaystyle\nabla . \textbf{v} = 0\quad\quad \quad\quad\quad\quad\quad\quad\quad\quad\\[8pt]
\mbox{\textbf{Momentum:} } \quad\quad\quad\quad \quad\quad\quad\displaystyle(\textbf{v}.\nabla)\textbf{v} + \nabla \textbf{p} -  \nabla.\left(\nu_t\nabla \textbf{\textbf{v}}\right) = 0\quad\quad\quad\quad\\[8pt]
\mbox{\textbf{TKE:} } \quad\quad\quad\quad\quad\quad\quad\quad\displaystyle(\textbf{v}.\nabla)\textbf{k} - \nabla .\left( \frac{\nu_t}{\sigma_k}\nabla\textbf{k}\right) - 2 \nu_t E.E + \textbf{e}  = 0\quad\quad\quad\quad\quad\quad\\[8pt]
\mbox{\textbf{TDR:} } \quad\quad\quad\quad\displaystyle(\textbf{v}.\nabla)\textbf{e} - \nabla .\left( \frac{\nu_t}{\sigma_e}\nabla\textbf{e}\right) - C_{1e} \frac{\textbf{e}}{\textbf{k}} 2 \nu_t E.E + C_{2e} \frac{\textbf{e}^2}{\textbf{k}}  = 0\quad\quad\quad\quad\quad\quad\\[8pt]
 \end{array}\right\}
\end{equation}

where, $k$ is Turbulent Kinetic Energy (TKE), $e$ is Turbulent Dissipation Rate (TDR), $v=(u_x, u_y, u_z)$ is the velocity vector, $p$ is pressure, $\nu_t$ is the kinematic turbulent viscosity, $C_{1e}, C_{2e}, \sigma_k, \sigma_e$ are empirical constants. More details related to this model and approach can be found in \citet{alfonsi2009reynolds}. Modeling of Reynolds averaged turbulence presents an additional challenge because it involves $2$ additional solution variables which are strongly coupled and influence the velocity and pressure of the flow.

The use case consists of a 3-D channel flow over arbitrarily shaped objects with a computational domain that has a grid resolution of $304$x$64$x$64$. The object is placed on the bottom surface of the channel to simulate conditions of automobiles moving on a road. The domain has a velocity inlet specified at $40$m/s and a zero pressure outlet boundary condition, while the rest of the surfaces are walls with no-slip conditions. The high resolution grid is required because the large flow Reynolds number, of the order of $10000$, results in large velocity gradients which can affect the entire flow field. In this experiment, we vary only the geometric features of the object but in future we would like to develop an external flow solver that encompasses both geometry and Reynolds number variations.

\subsubsection{Training} $150$ solutions are generated using Ansys Fluent for geometries corresponding to 5 primitive objects such as cylinder, cuboid, trapezoid, NURB and wedge and their random rotations along all axes as well as their combinations. The computational domain of $304$x$64$x$64$ is divided into smaller subdomains each with a resolution $8$x$8$x$8$. The solutions and geometry are encoded to a latent vectors of size $64$ and $32$, respectively. The geometry is represented with a signed distance field in each subdomain. The flux conservation network has a bottleneck layer of size $64$.

\subsubsection{Testing} $50$ additional solutions are generated using Ansys Fluent for other geometries sampled from the same distribution. These geometries are not a part of of the training set. The convergence tolerance of the CoMLSim solution algorithm is set to $1e^{-8}$ and the solution method is Gauss Seidel. Each solution converges in about $35$ seconds \color{black} and requires about 75 iterations on an average. \color{black}

\subsubsection{Comparison with Ansys Fluent for geometries outside the training set}


\begin{figure}[h!]
\centering
\includegraphics[width=0.65\linewidth]{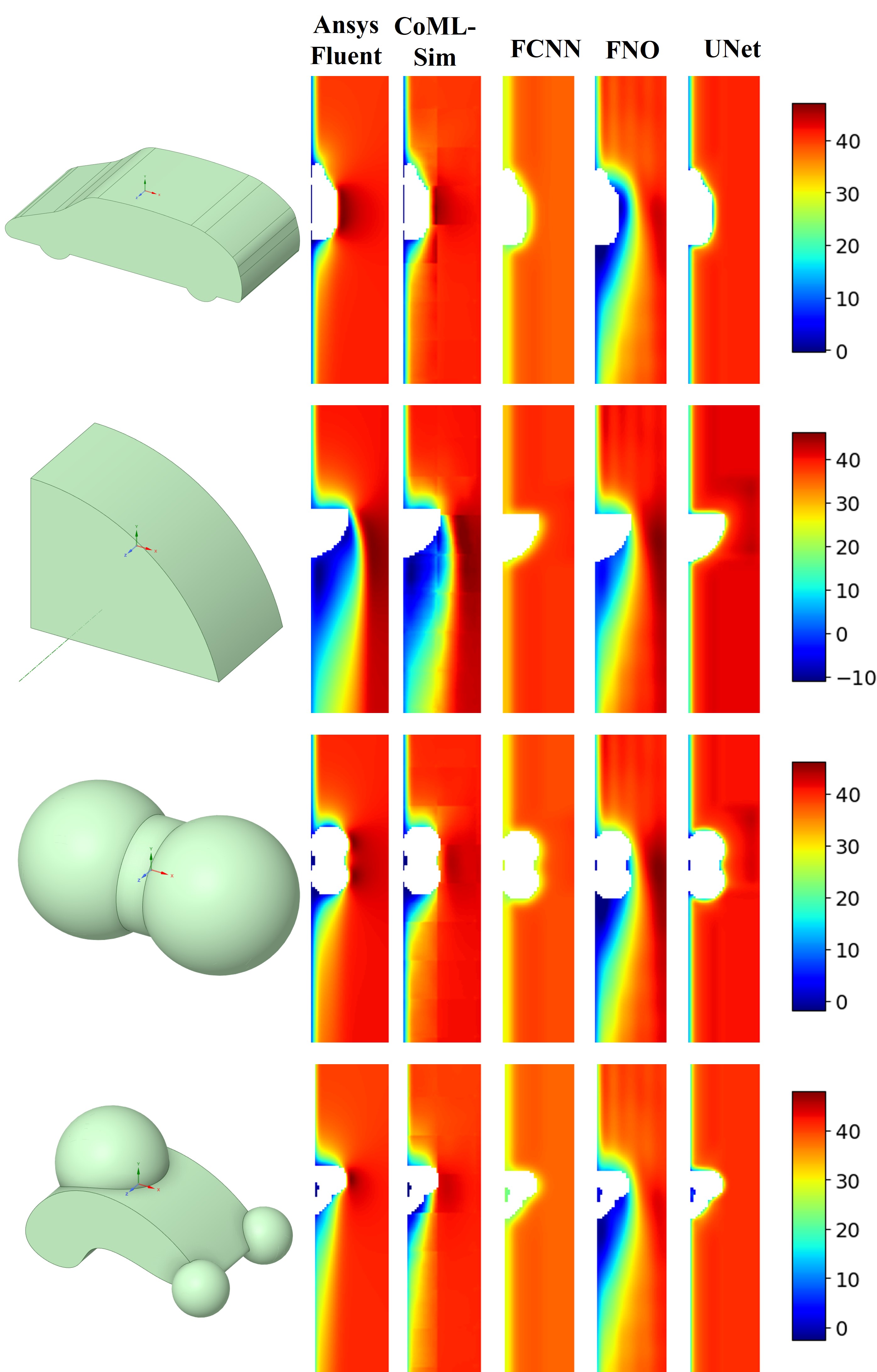}
\setlength{\belowcaptionskip}{-8pt}
\caption{X-velocity contour comparisons (Figures rotated to fit on page and zoomed in on the object)}
\label{fig:ex1}
\end{figure}

\begin{figure}[h!]
\centering
\includegraphics[width=0.65\linewidth]{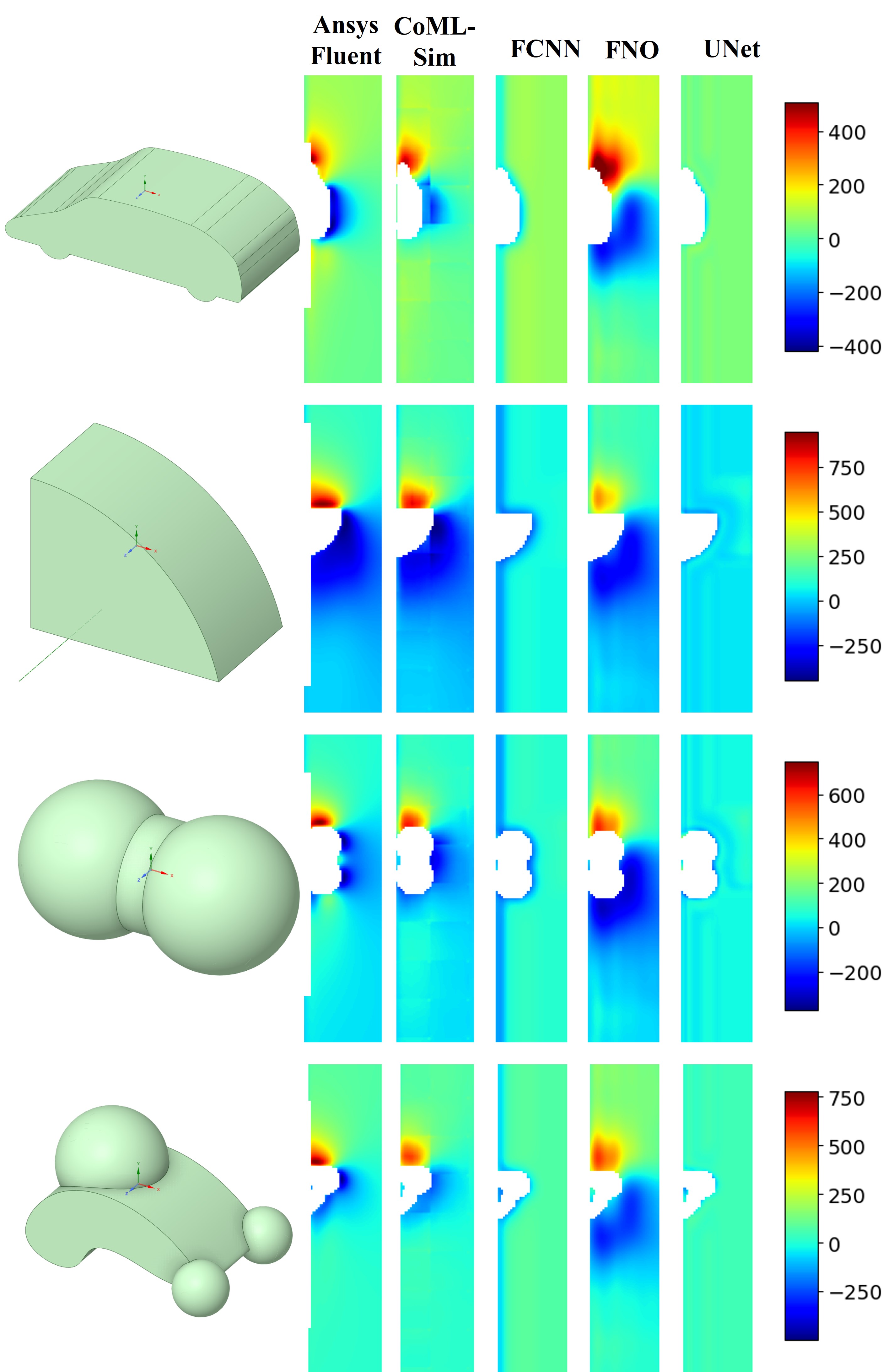}
\caption{Pressure contour comparisons (Figures rotated to fit on page and zoomed in on the object)}
\label{fig:ex2}
\end{figure}
It may be observed from Figures \ref{fig:ex1} and \ref{fig:ex2} that the CoMLSim predictions match well with Ansys Fluent for both x-velocity and pressure solutions. The geometries considered in this experiment were not a part of the training set and can be considered out-of-distribution for CoMLSim and other ML baselines. The different features of the flow field are captured reasonably well. As the flow first hits the object it severely decelerates causing smaller velocities and higher pressures. This is known as the stagnation point. Beyond this point, the flow travels along the surface of the object and eventually separates from the object leaving a recirculating wake on the downstream of the object. The point of separation is dependent on the geometry of the object and the flow speed, and determines the pattern of the recirculating wake. As the flow separates, it begins to accelerate resulting in a drop in pressure. It may be observed from the figure that all of these flow features are captured accurately by CoMLSim for the objects considered in this experiment. On the other hand, the other ML baselines do not capture the flow physics accurately. The error comparisons of CoMLSim and other ML baselines in comparison to Ansys Fluent are reported in the table \ref{tab:ext_flow}. It may be observed that CoMLSim outperforms the baselines.

\begin{table}
\centering
\caption{Mean Abs. Errors normalized by inlet velocity (40 m/s) in X-Velocity and Pressure}
    \begin{tabular}{|p{2cm}|p{2cm}|p{2cm}|}
\hline
&X-Velocity& Pressure\\
\hline
CoMLSim & 0.02025 & 0.088 \\
UNet & 0.0435 & 1.178 \\
FNO & 0.0315 & 0.5502 \\
FCNN & 0.1395 & 1.6685  \\
\hline
\end{tabular}
\label{tab:ext_flow}
\end{table}

\begin{figure}[h!]
\centering
\includegraphics[width=1\linewidth]{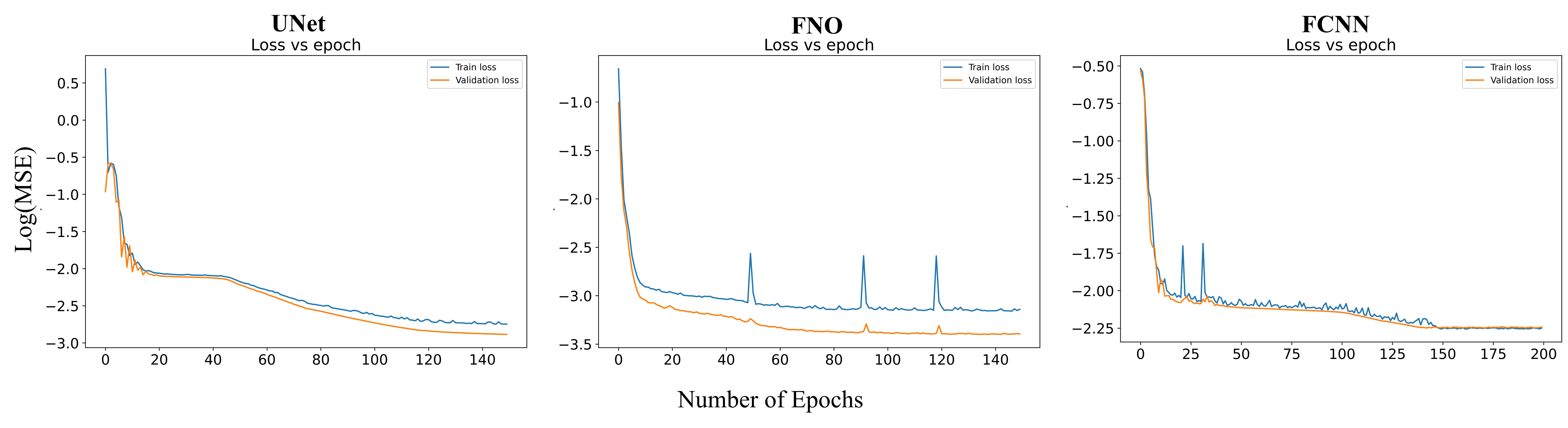}
\caption{Training loss curves for all the ML baselines}
\label{fig:ex3}
\end{figure}

Amongst the baselines, FNO performs the best and this is evident from the contours in Figures \ref{fig:ex1} and \ref{fig:ex2} as well as the errors reported in table \ref{tab:ext_flow}. On the other hand, FCNN and UNet are severely affected by the sparsity of the PDE solutions observed in all the flow variables and hence don't perform well. We would like to clarify that all the ML baselines were trained to the best of our ability. We provide loss curves in Figure \ref{fig:ex3} for the benefit of the reader to show that each baseline is trained until the training and validation losses plateau. Additionally, it may also be observed the loss drops by 3 orders of magnitude for each baseline. Moreover, as described in Section \ref{baseline_arch}, the networks have a sufficient number of parameters to train efficiently. Finally, traditional ML methods, such as the ones considered in this paper, are data-hungry and improve results as more data becomes available. However, in this experiment, we trained all the models with only $150$ solution samples on highly-resolved grids and this reflects in the poor solution accuracy of these methods. Increasing the depth of the networks and number of parameters to $5.38$ million for UNet and $2.33$ million for FCNN did not improve the results.

\color{black}
\subsubsection{Extension to larger physical domains and complex geometries}

In this section, we demonstrate the scalability of the CoMLSim approach to bigger physical domains with substantially higher mesh sizes. The results presented previously are carried out on physical domains with grid resolutions of 304x64x64. In this experiment, we increase the grid size to 562x128x128 by increasing the physical size of the domain as well as the geometry. Additionally, we use an unseen test geometry of a car model to perform this evaluation. 

\begin{figure}[h!]
\centering
\includegraphics[width=0.7\linewidth]{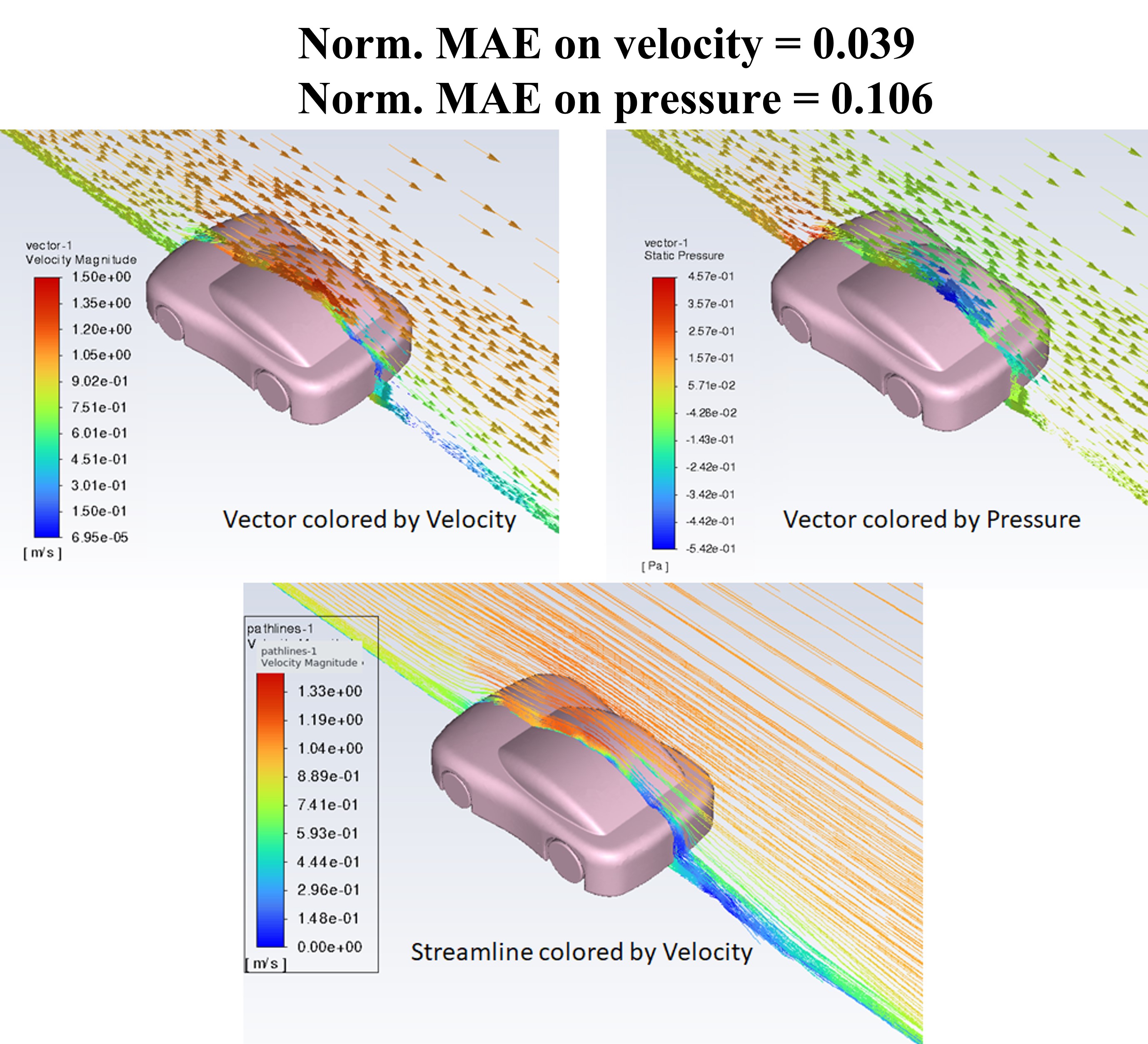}
\caption{CoMLSim evaluation on bigger domains and unseen car model}
\label{fig:ex4}
\end{figure}

Figure \ref{fig:ex4} shows the streamlines and vector plots of fluid flow near the surface of the car. When compared with Ansys Fluent we observe reasonable mean absolute errors of 0.039 for velocity and 0.106 for pressure. Moreover, it is important to note that the performance and scalability of our approach to larger meshes is not affected by the 3-D spatial representation in this use case. Most ML baselines considered in this work run out of computational memory when evaluating on this mesh.  

\color{black}
\subsection{Industrial usecase: 3-D electronic chip cooling with natural convection} \label{chip_cool}

In this section, we demonstrate the CoMLSim approach for solving the complicated industrial use case of electronic chip cooling with a wide range of applications. The main purpose of presenting this experiment is to show the ability of our approach can be used for solving real industrial application with similar accuracy's as traditional PDE solvers.

\begin{wrapfigure}{r}{0.35\textwidth}
    \centering
    \vspace{-1em}
    \includegraphics[width=0.35\textwidth]{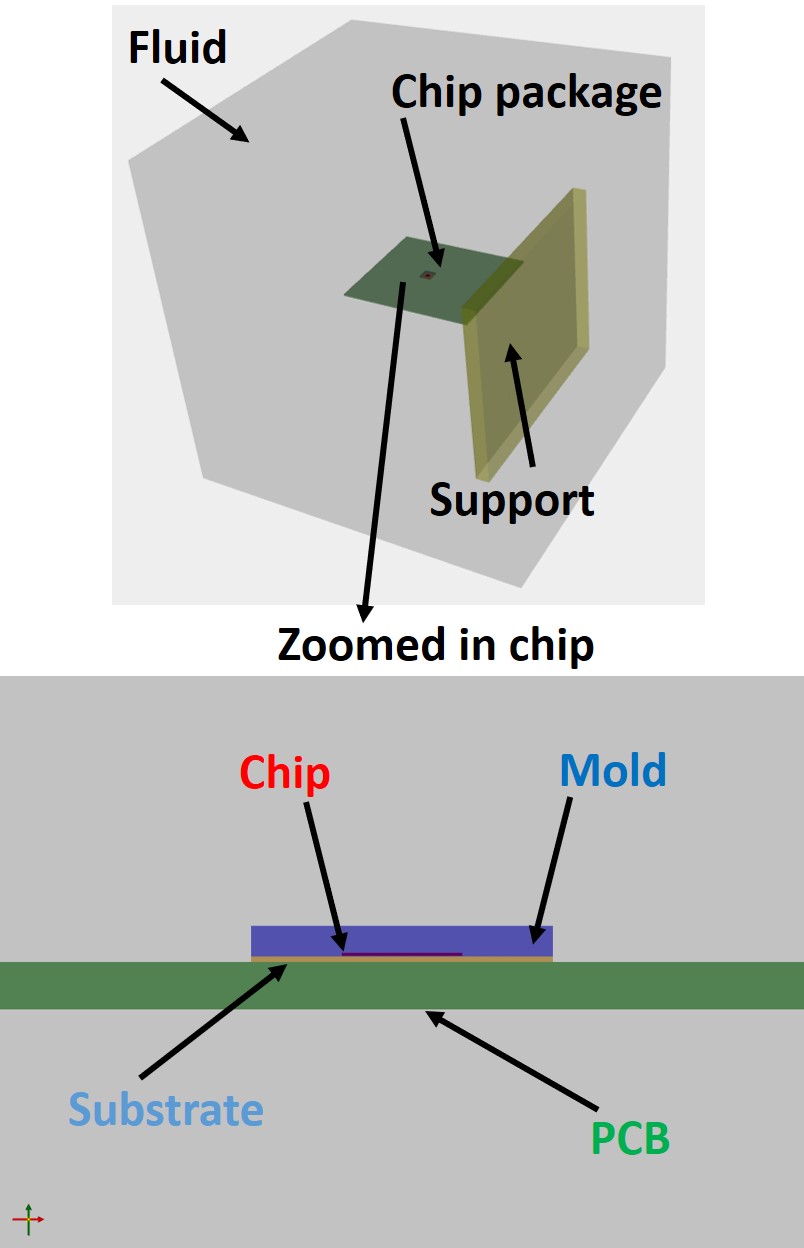}
    \vspace{-4ex}
    \setlength{\belowcaptionskip}{-88pt}
    \caption{Electronic chip cooling geometry}
    \label{chip_geo}
\end{wrapfigure}

In this case the domain consists of a chip, which is sandwiched in between an insulated mold. The chip-mold assembly is held by a PCB and the entire geometry is placed inside a fluid domain. The geometry and case setup of the electronic chip cooling case can be observed in Fig. \ref{chip_geo}. The chip is subjected to electric heating and the uncertainty in this process results in random spatial distribution of heat sources on the on the surface of the chip. The power map distribution in this case sampled from the same Gaussian mixture model described in Eq. \ref{eqc2}, but the total number of Gaussians is limited to $15$.

The physics in this problem is natural convection cooling where the power source is responsible for generating heat on the chip, resulting in an increase in chip temperature. The rising temperatures get diffused in to the fluid domain and increase the temperature of air. The air temperature induces velocity which in turn tries to cool the chip. At equilibrium, there is a balance between the chip temperature and velocity generated and both of these quantities reach a steady state. The objective of this problem is to solve for this steady state condition for an arbitrary power source sampled from a Gaussian mixture model distribution, which is extremely high dimensional. The governing PDEs that represent this problem are shown in Eqs. \ref{eqc6}.

\begin{equation}\label{eqc6}
\left. \begin{array}{ll}  
\mbox{\textbf{Continuity:} } \quad\quad\quad\quad\quad\displaystyle\nabla . \textbf{v} = 0\quad\quad \quad\quad\quad\quad\quad\quad\quad\quad\\
\mbox{\textbf{Momentum:} } \displaystyle(\textbf{v}.\nabla)\textbf{v} + \nabla \textbf{p} - \frac{1}{Re} \nabla^2 \textbf{\textbf{v}} + \frac{1}{\beta} \Vec{g} \textbf{T} = 0\quad\quad\quad\quad\\[8pt]
\mbox{\textbf{Heat (Solid):} } \quad\quad\quad\displaystyle\nabla .\left( \alpha\nabla\textbf{T}\right)-\color{black}P\color{black} = 0\quad\quad \quad\quad\quad\quad\quad\quad\\[8pt]
\mbox{\textbf{Energy (Fluid):} } \quad\displaystyle(\textbf{v}.\nabla)\textbf{T} - \nabla .\left( \alpha\nabla\textbf{T}\right) = 0\quad\quad\quad\quad\quad\\
 \end{array}\right\}
\end{equation}
where, $v={u_x, u_y, u_z}$ is the velocity field in $x, y, z$, $p$ is pressure, $T$ is temperature, $Re$ and $\alpha$ are flow and thermal properties, $P$ is the heat source term, $\frac{1}{\beta} \Vec{g} T$ is the buoyancy term. $P$ is the spatially varying power source applied on the chip center. The main challenges are in capturing the two-way coupling of velocity and temperature and generalizing over arbitrary spatial distribution of power.
The coupled PDEs with $5$ solution variables, $v={u_x, u_y, u_z}, P, T$ are solved on a fluid and solid domain with loose coupling at the boundaries. The fluid domain is discretized with $256^3$ elements in the domain and the solid domain (chip) is modeled as a 2-D domain with $64^2$ elements as it is very thin in the third spatial dimension. 

\subsubsection{Training} The data to train the autoencoders in the CoMLSim approach is generated using Ansys Fluent and corresponds to $300$ PDE solutions. The computational domain is divided into $4096$ subdomains, each with $16^3$ computational elements. The solution autoencoders for the $5$ solution variables are trained to establish lower dimensional latent vectors with size $29$ on the subdomain level.
\subsubsection{Testing} $100$ more solutions are generated for random Gaussian mixtures using Ansys Fluent. The convergence tolerance of the CoMLSim solution algorithm is set to $1e^{-8}$ and the solution method is Gauss Seidel. Each solution converges in about $42$ seconds on an average.

\subsubsection{Comparison with Ansys Fluent}

\textbf{Contour plots across different metrics:}

\begin{figure}[h!]
\centering
\includegraphics[width=0.6\linewidth]{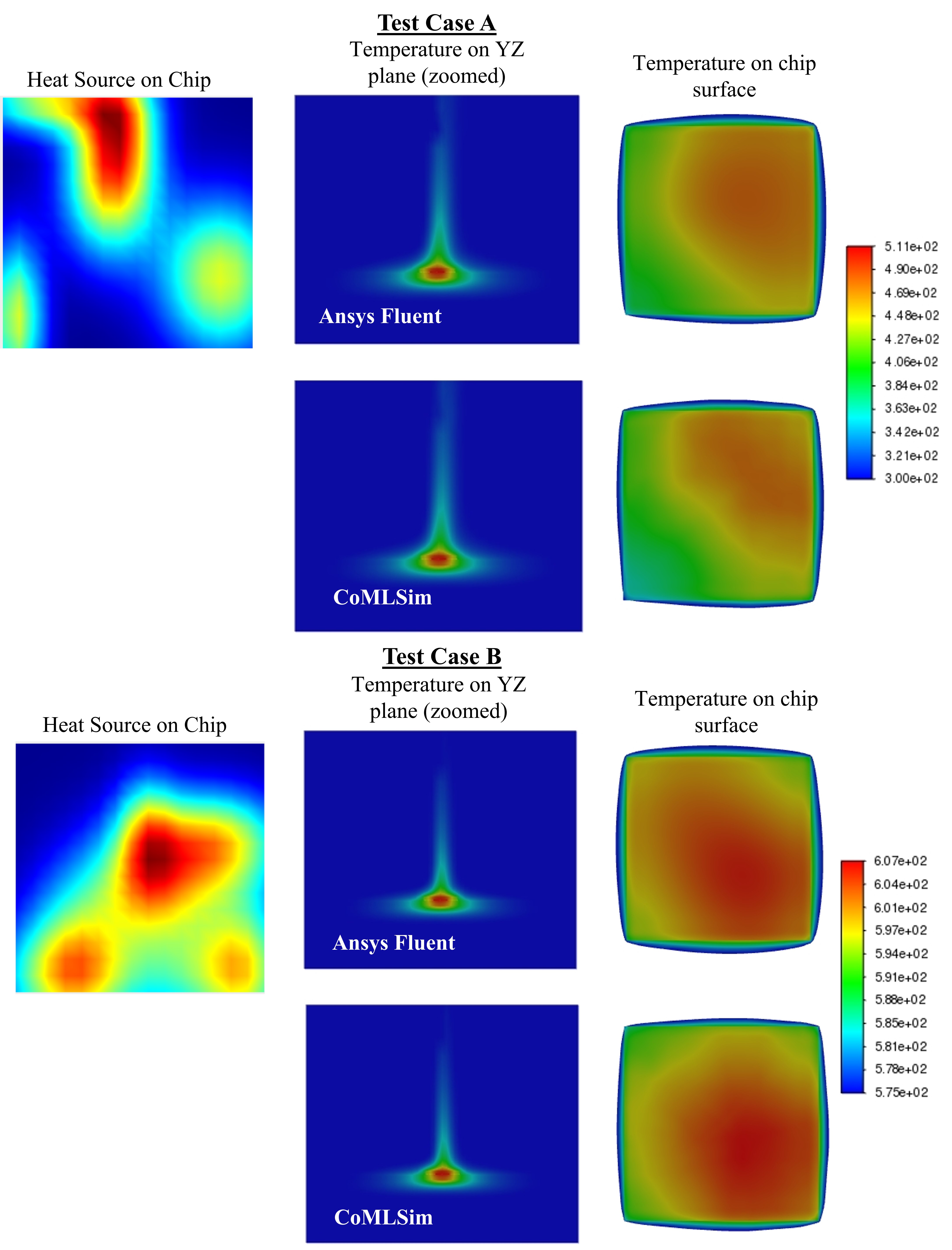}
\caption{Temperature contour comparisons for unseen powermaps}
\label{fig:ch1}
\end{figure}

\begin{figure}[h!]
\centering
\includegraphics[width=0.6\linewidth]{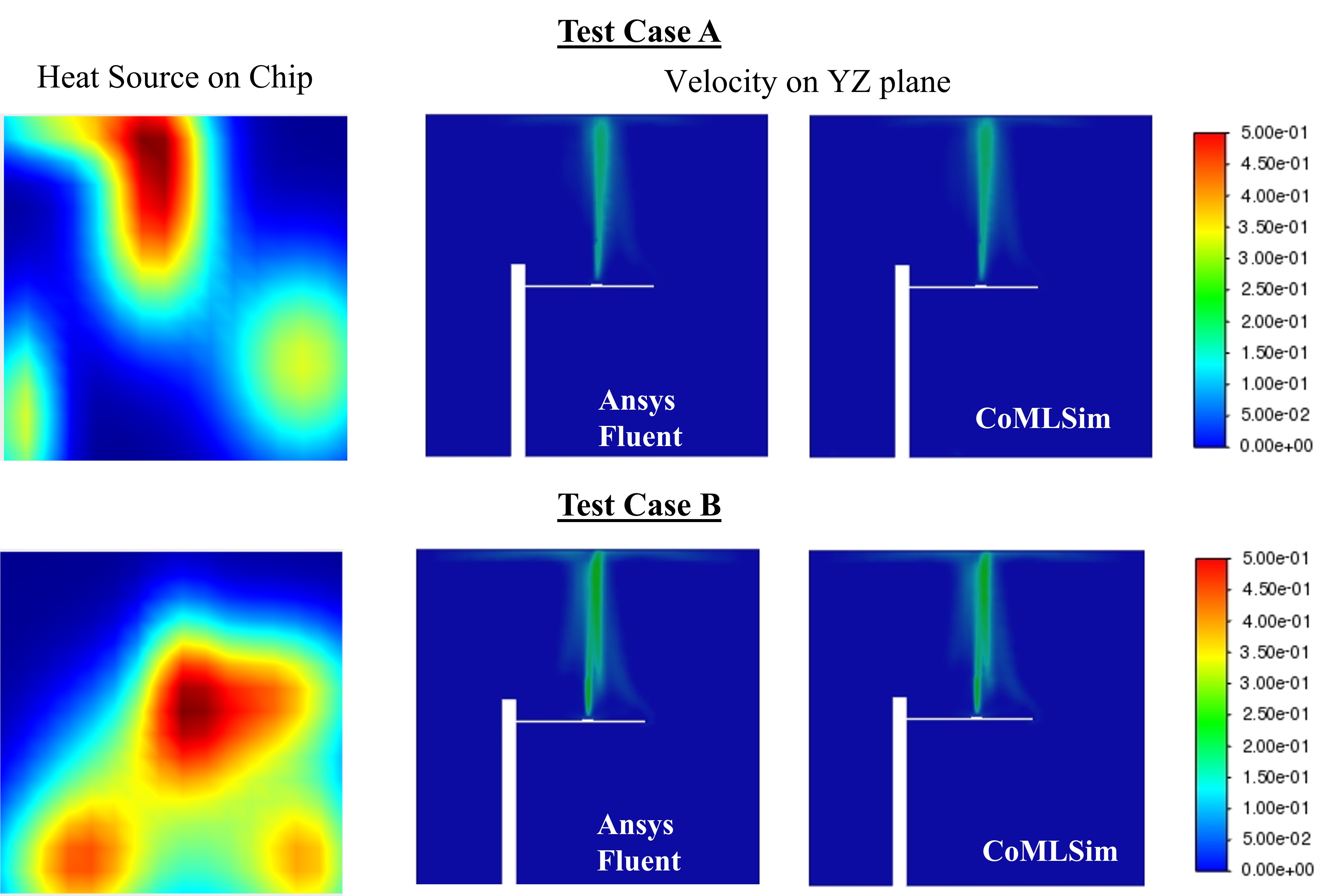}
\caption{Velocity contour comparisons for unseen powermaps}
\label{fig:ch2}
\end{figure}

The CoMLSim predictions for $2$ randomly selected unseen testing samples are compared with Ansys Fluent in Figs. \ref{fig:ch1} and \ref{fig:ch2}. It may be observed that the contour comparisons agree well with Fluent solutions and the mean absolute errors are of the order of $1e^{-2}$ for temperature and $1e^{-3}$ for velocity. Similar results are observed for other power maps from the unseen testing set.

\textbf{Line plots across different metrics:}

\begin{figure}[h!]
\centering
\includegraphics[width=0.6\linewidth]{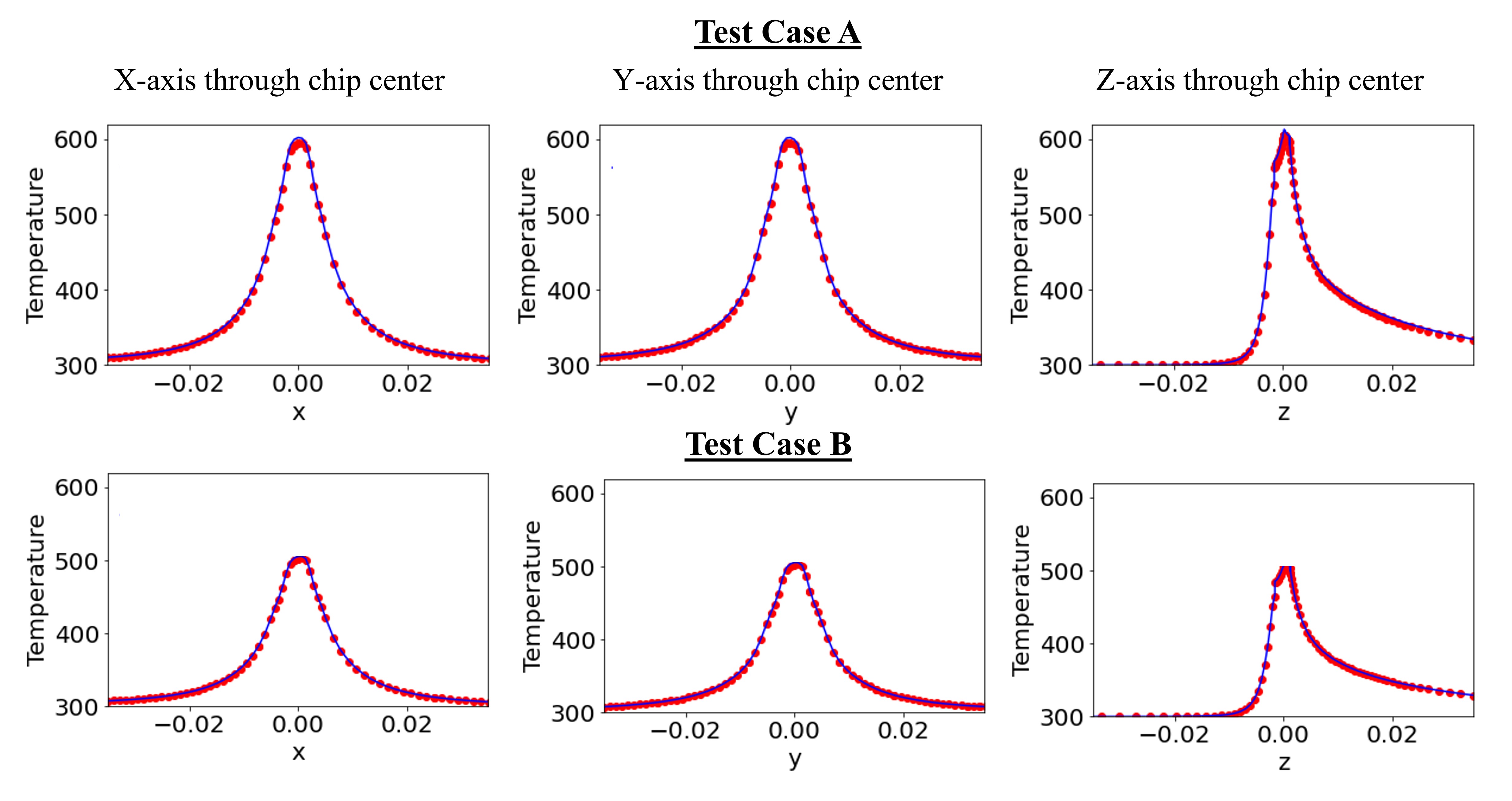}
\caption{Line comparisons of temperature for unseen powermaps (\color{black}blue\color{black}:CoMLSim, \color{red}red\color{black}:Ansys Fluent}
\label{fig:ch4}
\end{figure}

\begin{figure}[h!]
\centering
\includegraphics[width=0.6\linewidth]{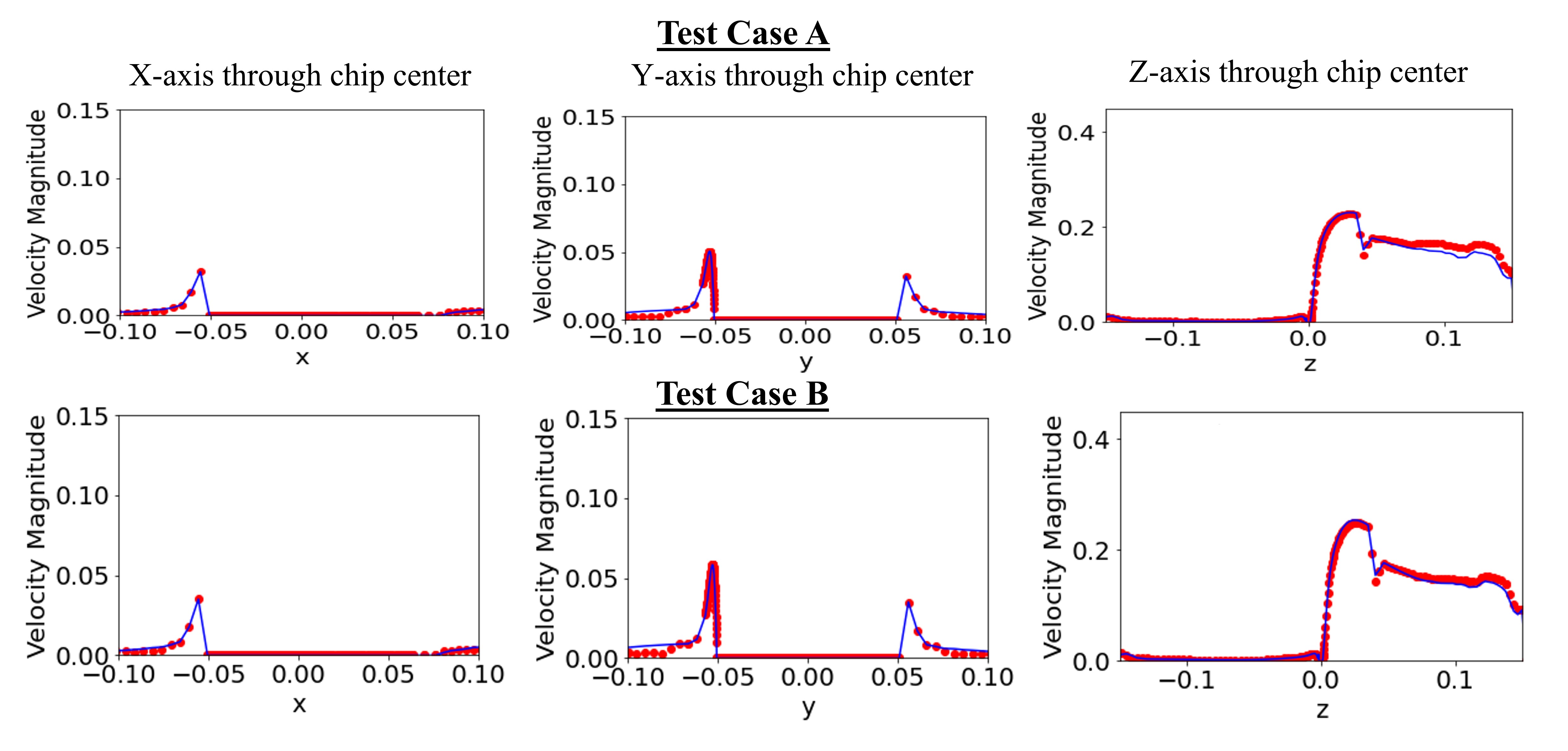}
\caption{Line comparisons of velocity for unseen powermaps (\color{black}blue\color{black}:CoMLSim, \color{red}red\color{black}:Ansys Fluent}
\label{fig:ch5}
\end{figure}

Figures \ref{fig:ch4} and \ref{fig:ch5} shows the line comparisons of temperature and velocity plotted along the chip across all three spatial directions. These metrics provide a more quantitative representation of the accuracy of our approach. It maybe observed that the predictions of CoMLSim agree well with Ansys Fluent for the 2 test cases.

\textbf{Comparisons with ML baselines for other quantitative metrics:}

\begin{table}[h!]
\centering
\caption{Testing errors for different metrics}
    \begin{tabular}{|p{2cm}|p{2cm}|p{2cm}|p{3cm}|}
\hline
&$L_{\infty}(T)$& $\epsilon(T_{max})$&$\ \epsilon(heat\ flux)$\\
\hline
CoMLSim & 15.2 & 5.09 & 1.26\\
UNet & 95.21 & 76.2 & 27.56 \\
FNO & 60.83 & 5.97 & 12.12 \\
FCNN & 192.7 & 191.38 & 311.23 \\
\hline
\end{tabular}
\label{tab:comparison_unet_nn}
\end{table}

Finally, in Table \ref{tab:comparison_unet_nn} we present $3$ different testing metrics for the CoMLSim predictions, 1) Error in maximum temperature in computational domain (hot spots on chip), 2) $L_{\infty}$ error in temperature in the computational domain, and 3) Error in heat flux (temperature gradient) on the chip surface. These metrics are more suited for this application and provide a much better measure for evaluating accuracy. All errors are computed relative to the Ansys Fluent solution and are averaged over $100$ the unseen testing samples. It may be observed from Table \ref{tab:comparison_unet_nn} that CoMLSim agrees well with Ansys Fluent on all the comparison metrics and outperforms all the ML baselines. 

\section{Experiments for additional canonical PDEs} \label{expt_additional}

\subsection{2-D Laplace Equations:}

The Laplace equation represents a canonical problem for benchmarking linear solvers. It is shown below in Eq. \ref{eqd1}.

\begin{equation}
        \nabla^2 \phi = 0
    \label{eqd1}
\end{equation}

where, $u$ is the solution variable subjected to a Dirichlet boundary condition, $\phi (\vec{x_b}) = f_b$ or a Neumann boundary condition, $\frac{\partial \phi}{\partial \vec{x_b}} = f_b$. The boundary conditions (BCs) are sampled randomly and the magnitude of the BC is uniformly sampled between $-1.0$ and $1.0$.

\subsubsection{Training} $300$ solutions are generated for arbitrary BCs using Ansys Fluent on a computational domain with a resolution of $256$x$256$. These solutions are used to train the different components of CoMLSim. The computational domain is divided into $64$ subdomains each of resolution $32$x$32$. The solution is compressed into latent vectors of size $7$. Since the boundary condition is uniform, it is encoded discretely on each subdomain. The boundary condition encoding is a vector of length $8$, such that the first $4$ elements represent an index, $0$ or $1$ indicating Dirichlet or Neumann boundary on the $4$ sides of a subdomain. The next $4$ elements represent the magnitude of BCs on the respective sides. The flux conservation autoencoder has a bottleneck layer of size $35$. 

\subsubsection{Testing} $100$ more solutions are generated for arbitrary BCs using Ansys Fluent. The testing set contains completely different boundary conditions than the training set with no overlaps. The convergence tolerance of the CoMLSim solution algorithm is set to $1e^{-8}$ and the solution method is Point Jacobi.

\subsubsection{Comparisons with Ansys Fluent}
\begin{figure}[h]
\centering
\includegraphics[width=0.5\linewidth]{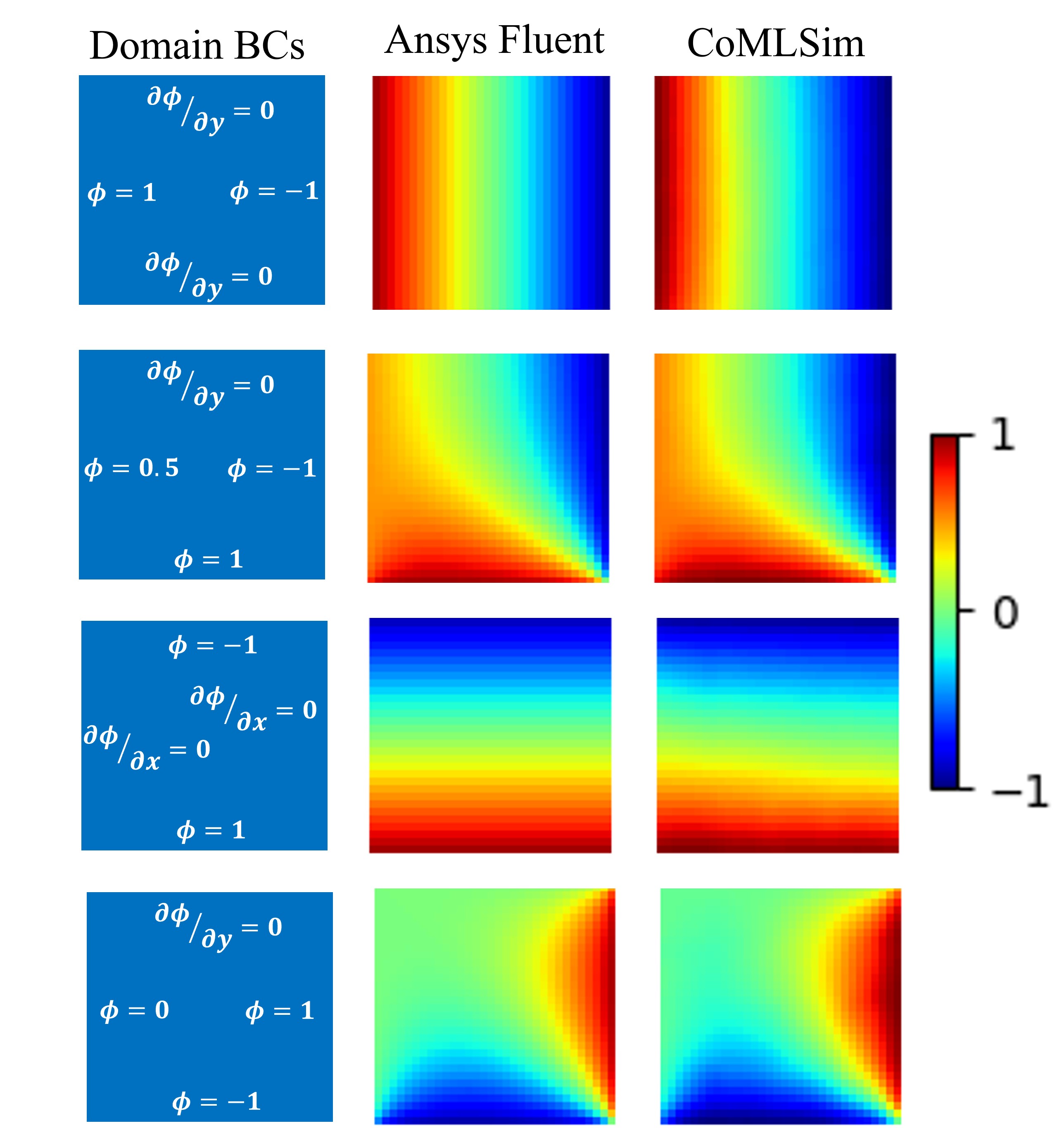}
\caption{CoMLSim vs Ansys Fluent for Laplace equation over different unseen BCs}
\label{fig:5a}
\end{figure}
The CoMLSim predictions for a selected unseen testing samples are compared with Ansys Fluent in Fig. \ref{fig:5a}. It may be observed that the contour comparisons agree well with Fluent solutions. Overall, the mean absolute error over $100$ testing samples is $0.015$.

\subsection{2-D Darcy Equations:}

The Darcy equation is defined as follows:
\begin{equation} \label{darcy_eq}
    -\nabla.(\alpha\nabla\phi (\vec{x}) = f
\end{equation}

It is subjected to different diffusivity conditions, $\alpha(\vec{x}) = f(\vec{x})$ in a 2-D bounded computational domain between (0, 1). The problem setup as well as the data to train the autoencoders is taken from \cite{li2020fourier}. The main objective for including this case was to demonstrate the CoMLSim approach on a public data set and on relatively smaller mesh resolutions.

We use about $400$ samples for training the flux conservation autoencoders, and $200$ randomly picked samples for testing from the remaining data. The computational domain in this problem is 2D and has a resolution of $241$x$241$. The domain is divided into $241$ subdomains of resolution $21$x$21$ each. The solution on each subdomain is encoded into a latent vector of size $11$ and the diffusivity is encoded into size $11$. The flux conservation network has a bottleneck size of $16$. Since, this is a steady state problem, the CoMLSim iterative solution algorithm is initialized with a solution field equal to zero in all test cases. The iterative algorithm convergence tolerance is set to $1e^{-7}$. 

The testing error is computed using Eq. \ref{test_error} shown below:
\begin{equation}\label{test_error}
    \epsilon = \frac{L_2 (Y_{pred}-Y_{true})}{L_2(Y_{true})}
\end{equation}

The CoMLSim has a testing error of $0.019$ averaged over $200$ unseen testing samples as compared to a second-order finite difference solver. The CoMLSim error compares well with FNO \cite{li2020fourier}, which has an error of $0.021$. Although the testing errors are very similar in this case, the main advantage of the CoMLSim approach is to accurately model local solution features on highly resolved grids and out-of-distribution generalizability as presented in the different use cases in the main paper as well as the supplementary materials.

\section{Computational Speed Analysis:} \label{comp_perf}

We observe that the CoMLSim approach is about $40$-$50$x faster as compared to commercial steady-state PDE solvers such as Ansys Fluent for the same mesh resolution in all the experiments presented in this work. Both CoMLSim approach and Ansys Fluent are solved on a single Xeon CPU with single precision. We expect our approach to scale to multiple CPUs as traditional PDE solvers but single CPU comparisons are provided here for benchmarking. Moreover, our algorithm is a python language interpreted code, whereas Ansys Fluent is an optimized, C language pre-compiled code. We expect the C/C++ version of our algorithm to further provide independent speedups (not included in current estimates). In Table \ref{tab:time_baseline}, we show comparison of simulation time between Ansys Fluent and CoMLSim for all the use cases. The results are averaged over $100$ testing cases. As expected, the iterative convergence achieved by the CoMLSim approach is substantially faster than Ansys Fluent. 

In comparison to the ML baselines, our approach is expected to be slower because it adopts an iterative inferencing approach. But, our approach focuses on substantially improving the solution accuracy, generalizability, robustness and stability at the cost of computational speed-ups. Although our approach is still faster than traditional PDE solvers, but the ratio of is speed-up is smaller than traditional ML methods.

\begin{table}[h]
\caption{Simulation time (in seconds) comparison with Ansys Fluent}
\centering
\begin{tabular}{ |p{4cm}|p{2cm}|p{2cm}|p{3cm}|  }
\hline
Experiment& Num. of Elements &CoMLSim& Ansys Fluent\\
\hline
Laplace & 65K & 0.21 & 10\\
Linear Poissons   & 1048K   & 2.75&   130 \\
Coupled non linear Poissons &   1048K & 2.81 & 540 \\
3D Navier-Stokes flow &   1245K & 35 & 1900 \\
3D Electronic Chip Cooling &   2097K & 42 & 1600 \\
\hline
\end{tabular}
\label{tab:time_baseline}
\end{table} 

\section{Guidelines for reproducibility} \label{reproduce}

\begin{figure}[h!]
 \centering
  \includegraphics[width=\textwidth]{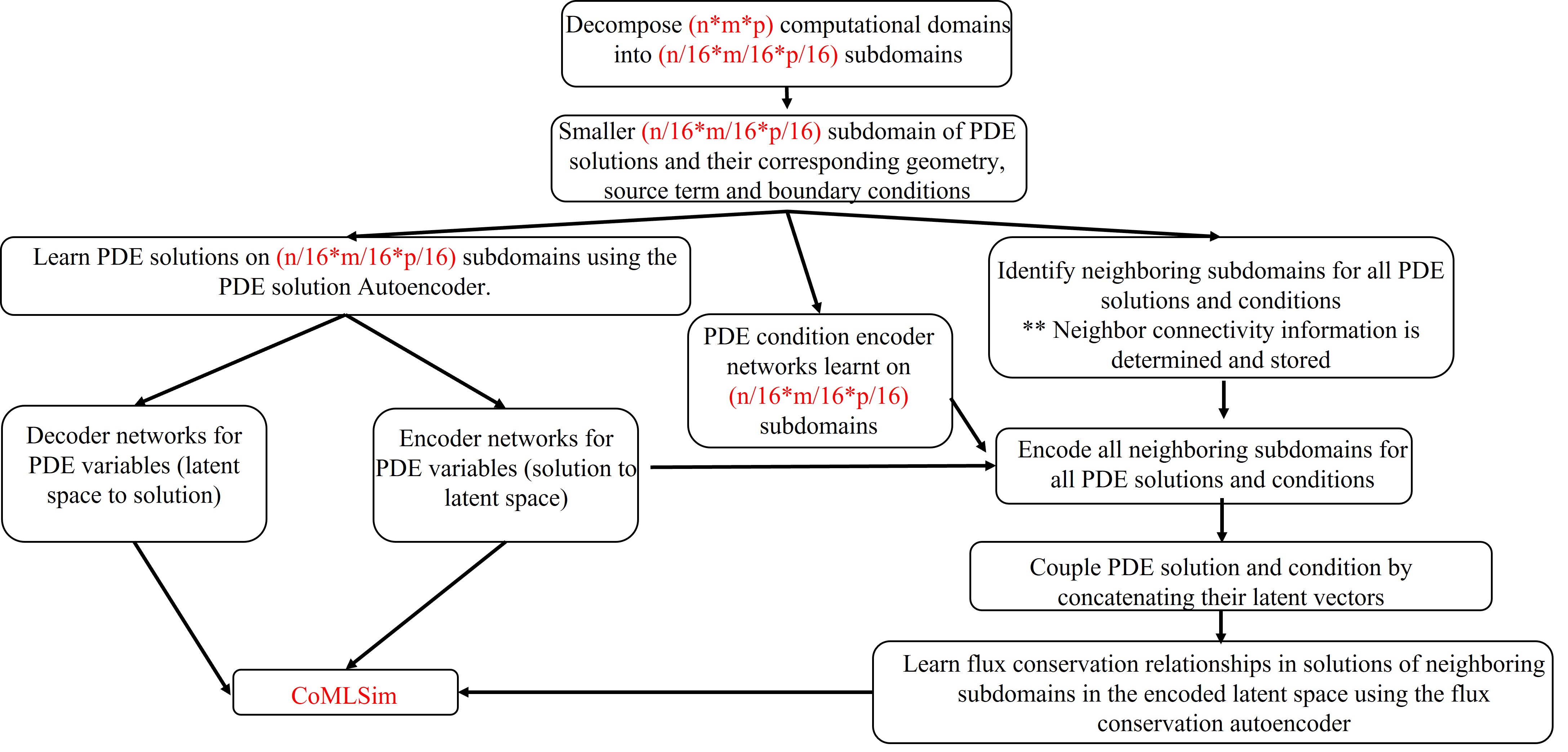}
  \caption{Flow chart for training the CoMLSim approach}
  \label{mlsolver_reproduce}
\end{figure}

In this section, we provide the necessary details for training the CoMLSim. As emphasized previously, the CoMLSim training corresponds to training several autoencoders for PDE solutions, conditions, such as geometry, boundary conditions and source terms and for flux conservation. In Figure \ref{mlsolver_reproduce}, we present a flow chart of the steps that can be followed to train each of these autoencoders. The specific training details and network architectures are described in the previous Section. 
For a given set of coupled PDEs and depending of the use case, we start with $100$-$1000$ sample solutions on a computational domain with $n, m, p$ computational elements in spatial directions $x, y, z$, respectively. The solutions are divided into smaller subdomains of resolution, $n/16, m/16, p/16$. Each subdomain can has PDEs solutions and PDE conditions associated with it. The PDE solutions are used as training samples to the PDE solution autoencoder to learn a compressed encoding of all the variables on the subdomain. On the other hand, the autoencoders for PDE conditions can be trained with completely random samples, which may or may not be related to sample solutions. Once the PDE solution and condition autoencoders are trained, neighboring subdomains are grouped together and the solution and PDE condition latent vectors on groups of neighboring subdomains are stacked together. These groups of stacked latent vectors are used for training the flux conservation autoencoder. The trained PDE solution, condition and flux conservation autoencoders combine to form the primary components of the CoMLSim. 

The solution algorithm of the CoMLSim has been described in detail in the main body of the paper and may be used for solving for unseen PDE conditions. We are happy to share the source code and the dataset for some experiments if deemed necessary by the reviewers.

\color{black}
\subsection{Step-by-step instructions for Poisson's equation}

The Poisson's equation is shown in Eq. \ref{eqc1}. The solution to this equation is governed by a source term applied on the computational domain. Here we consider this 2D case to explain the training and inference modes of the CoMLSim approach. 

\subsubsection{CoMLSim training}

The data in this use case corresponds to $256$ training and testing PDE solutions generated for random source terms defined by Eq. \ref{eqc2} using Ansys Fluent on a 1024x1024 resolution grid. This data is preprocessed such that it is divided into subdomains of size 64x64 before being used for training. 

The training portion of CoMLSim corresponds to training the solution autoencoder, source term autoencoder and the flux conservation autoencoder. Training solutions and the corresponding conditions on 64x64 subdomains are used for training the solution and condition autoencoders. The Network architecture and training mechanics described in Figure \ref{solution_encoder} are used to perform this training. Separate 2D CNN autoencoders are used to train solutions and source terms.

The flux conservation networks require further preprocessing of the data before it is trainable. The trained encoders are used to encode the solutions and conditions on each subdomain and its top, bottom, left and right neighbors. These neighborhood encodings are concatenated. For example, if the solution latent size is 16 and the condition latent size is 16, then the input size for the flux conservation autoencoder is (16+16)*5=160. The flux conservation autoencoder is trained to learn this group of concatenated neighborhood encodings for all the subdomains in the training dataset. For the subdomains on the edges, we simply use zero vectors to pad the missing subdomains. For 3-D cases, we include two additional subdomains, one in front and behind the center subdomain. The network architecture and training mechanics used to perform this training is described in more detail in Figure \ref{flux_encoder}. 

The trained autoencoders are used to evaluate the our approach on other unseen source terms.

\subsubsection{CoMLSim inference}

\begin{figure}[h!]
 \centering
  \includegraphics[width=\textwidth]{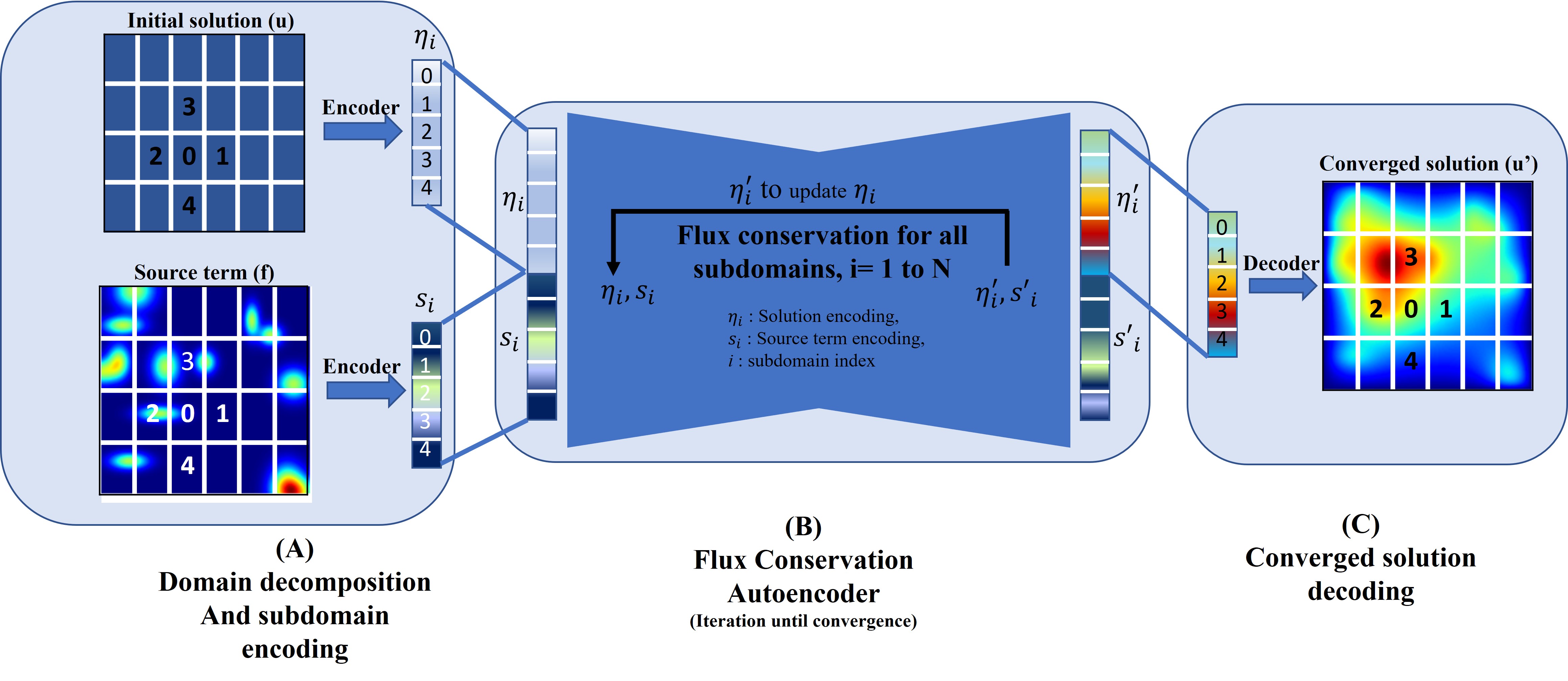}
  \caption{Schematic of CoMLSim inference for Poisson's PDE: The inference process consists of 3 steps. (A) Domain decomposition and encoding: The solution and source term are divided into subdomains. The central subdomain and the neighboring subdomains are encoded individually using the trained autoencoders. $\eta_i$ represents the encoded initial solutions and $s_i$ the encoded subdomains. The solution and source latent vectors on neighboring subdomains are concatenated together. (B) Flux conservation: The concatenated encodings are passed through the flux conservation autoencoder iteratively until convergence. The source term $s_i$, part of the encoding is held constant. Only the solution $\eta_i$, part is updated. In each flux conservation iteration,  the algorithm loops over all the subdomains before moving to the next iteration. (C) The converged solution encoding is then extracted from the complete encoding and decoded per subdomain using the solution decoder trained during the training phase.
}
  \label{mlsolver_example}
\end{figure}

During the inference time, we evaluate the CoMLSim approach for a user-specified source term. A schematic of the inference algorithm for this case is shown in Figure. \ref{mlsolver_example}. 

Similar to training, we decompose the domain into same sized smaller subdomains and encode the solution and source terms on all the subdomains using the pretrained encoders. Since the solution is unknown, a constant or randomly initialized solution is assumed on the domain. In each fixed point iteration, we loop over all the subdomains. On each subdomain, we gather the solution and source term latent vectors of the subdomain and its neighbors. The order in which the neighboring latent vectors are concatenated is consistent with the order used during training. The concatenated neighborhood latent vectors are then passed through the flux conservation autoencoder to obtain a new set of solution and source term encodings for each subdomain and its neighbors. Depending on the algorithm being used, Point Jacobi or Gauss Seidel, we modify the solution encoding of the subdomains while maintaining the source term encoding fixed. The modified solution encodings along with the fixed source term encodings are iteratively passed through the flux conservation autoencoder until the L-2 norm of the change in the solution latent vectors on all subdomains drops below a specified tolerance. The constant source term encodings steer the solution encoding towards an equilibrium. Finally, the converged solution encodings are decoded using the trained solution decoder. The decoded solutions are then stitched together to obtain the full domain solution.

\color{black}



\end{document}